# Robust Wirtinger Flow for Phase Retrieval with Arbitrary Corruption

Jinghui Chen[*] and Lingxiao Wang[†] and Xiao Zhang[‡] and Quanquan Gu[§]

## Abstract

We consider the robust phase retrieval problem of recovering the unknown signal from the magnitude-only measurements, where the measurements can be contaminated by both sparse arbitrary corruption and bounded random noise. We propose a new nonconvex algorithm for robust phase retrieval, namely Robust Wirtinger Flow to jointly estimate the unknown signal and the sparse corruption. We show that our proposed algorithm is guaranteed to converge linearly to the unknown true signal up to a minimax optimal statistical precision in such a challenging setting. Compared with existing robust phase retrieval methods, we achieve an optimal sample complexity of $O(n)$ in both noisy and noise-free settings. Thorough experiments on both synthetic and real datasets corroborate our theory.

## 1 Introduction

In the fields of machine learning, signal processing and statistics, one important problem is to solve a quadratic system of equations. Specifically, we are interested in solving the following system of $m$ quadratic equations:

$$y_i = |\langle \mathbf{a}_i, \mathbf{x}^* \rangle|, \ 1 \leq i \leq m, \tag{1.1}$$

where $\mathbf{x}^* \in \mathbb{R}^n$ or $\mathbb{C}^n$ is the unknown signal we try to recover. $\mathbf{a}_i \in \mathbb{R}^n$ or $\mathbb{C}^n$ is the design/sensing vectors and $\mathbf{y} = (y_1, y_2, \ldots, y_m)^\top$ is the observation vector. Equivalently, (1.1) can be written as quadratic form: $y_i^2 = |\langle \mathbf{a}_i, \mathbf{x}^* \rangle|^2$. Due to its combinatorial nature caused by the missing signs of $\langle \mathbf{a}_i, \mathbf{x}^* \rangle$, solving such a quadratic system of equations is generally considered as NP-hard (Pardalos and Vavasis, 1991).

In the literature of physical sciences, the problem of solving (1.1) is also known as phase retrieval (Fienup, 1978; Candes et al., 2015b), where the goal is to reconstruct the unknown signal vector from magnitude only measurements. There exists a large body of literature (Fienup, 1978, 1982;

---

[*]Department of Computer Science, University of Virginia, Charlottesville, VA 22904, USA; e-mail: jc4zg@virginia.edu

[†]Department of Computer Science, University of Virginia, Charlottesville, VA 22904, USA; e-mail: lw4wr@virginia.edu

[‡]Department of Computer Science, University of Virginia, Charlottesville, VA 22904, USA; e-mail: xz7bc@virginia.edu

[§]Department of Computer Science, University of Virginia, Charlottesville, VA 22904, USA; e-mail: qg5w@virginia.edu

Gerchberg, 1972; Candes et al., 2013, 2015a; Netrapalli et al., 2013; Candes et al., 2015b; Goldfarb and Qin, 2014; Wei, 2015; Zhang and Liang, 2016; Wang et al., 2016b; Sun et al., 2016; Wang et al., 2016a; Zhang et al., 2016b; Huang et al., 2016; Goldstein and Studer, 2017) for phase retrieval in the noise-free and noisy cases. The applications of phase retrieval include X-ray crystallography (Harrison, 1993; Miao et al., 1999), microscopy (Miao et al., 2008), diffraction and array imaging (Bunk et al., 2007; Chai et al., 2010), optics (Millane, 1990) and so on.

In many applications, it is not uncommon that the measurements $|\langle \mathbf{a}_i, \mathbf{x}^* \rangle|$'s are corrupted by errors[1]. The corruption arises due to various reasons such as illumination, occlusion, device malfunctioning, damage of measuring equipment or simply recording errors. These types of corruption are usually large in magnitudes and do not disappear by averaging the results. Hence it is of great importance for the phase retrieval algorithms to be able to handle these corruption that can be arbitrarily large, and if possible, identify the location of the corruption. Nevertheless, most of existing phase retrieval algorithms do not have an intrinsic mechanism to deal with arbitrary corruption, and would fail when the corruption is present.

In this paper, we study the robust phase retrieval problem. More specifically, we aim to develop a new phase retrieval algorithm that is able to recover the unknown signal from arbitrary large sparse corruption. Our work is along the line of the Wirtinger flow (WF)-type approaches (Candes et al., 2015b; Chen and Candes, 2015; Zhang and Liang, 2016; Wang et al., 2016b), which solves the problem by minimizing a nonconvex loss function with gradient descent algorithm and can be shown to converge to the unknown signal under good initialization. By using the "reshaped" amplitude-based loss function (Zhang and Liang, 2016; Wang et al., 2016b), our proposed robust Wirtinger Flow (Robust-WF) algorithm is proved to be robust against arbitrarily large corruption. Experiments on both synthetic data and real data verify the advantages of our algorithm and corroborate our theory. The main contributions of this paper are highlighted as follows:

- Unlike existing algorithms (Zhang et al., 2016a; Hand and Voroninski, 2016) which only estimate the unknown signal, our proposed Robust-WF algorithm jointly estimates the unknown signal and the sparse corruption without discarding any given information.

- Our proposed algorithm is guaranteed to converge linearly to the unknown true signal up to a minimax optimal statistical precision in noisy setting and exactly recover the unknown signal in the noise-free setting. More importantly, our algorithm achieves the optimal $O(n)$ sample complexity in both settings, which improves the previous robust phase retrieval results (Zhang et al., 2016a; Hand and Voroninski, 2016).

- Last but not least, the computational complexity of our algorithm is $O(mn \cdot \log(1/\epsilon))$, which matches the state-of-the-art. In other words, our algorithm is able to recover the unknown signal from corrupted measurements without paying any additional computational price.

**Notation.** For a vector $\mathbf{x} \in \mathbb{R}^n$, define vector norm as $\|\mathbf{x}\|_2 = \sqrt{\sum_{i=1}^n x_i^2}$, the infinity norm as $\|\mathbf{x}\|_\infty = \max_i \{x_i\}$. And $\|\mathbf{x}\|_0 = \sum_{i=1}^n \mathbb{1}\{x_i \neq 0\}$ denotes the number of nonzero entries in $\mathbf{x}$. For a matrix $\mathbf{A} \in \mathbb{R}^{m_1 \times m_2}$, we denote the spectral norm $\|\mathbf{A}\|_2 = \max_{\|\mathbf{u}\|_2=1} \|\mathbf{A}\mathbf{u}\|_2$. For two sets $S_1$ and $S_2$, we denote $S_1 \setminus S_2 = \{x \in S_1, x \notin S_2\}$ as the relative complement set and $S_1^c$ as the complement set of $S_1$. Further we denote the sign function as $\text{sgn}(t) = t/|t|$ indicators the sign of $t$.

**Organization.** The remainder of this paper is organized as follows: in Section 3, we review the problem formulation in detail. We present the algorithm in Section 4, and the main theory in Section

---

[1] It is important to distinguish corruption from random noise.

5. In Section 6, we sketch the proof of the main theory. In Section 7, we compare the proposed algorithm with existing algorithms on both synthetic data and real-world datasets. Finally, we conclude this paper in Section 8.

## 2 Related Work

Various techniques have been developed for solving the phase retrieval problem. They can be generally classified into two categories: convex approaches and nonconvex approaches. Convex methods like PhaseLift (Candes et al., 2013), PhaseCut (Waldspurger et al., 2015) adopt a so-called matrix-lifting technique to linearize the constraint by introducing a rank-one matrix and then relax the rank-one condition. Recently another convex method called PhaseMax (Goldstein and Studer, 2017) was proposed, which operates in the original signal space rather than lifting it to a higher dimensional space. While convex approaches do enjoy good recovery guarantees, their computational complexities are usually too large to afford especially when the dimension of the signal is high.

On the other hand, nonconvex approaches including Gerchberg-Saxton (Gerchberg, 1972), Fienup (Fienup, 1982), AltMinPhase (Netrapalli et al., 2013), trust-region (Sun et al., 2016), choose to directly optimize the nonconvex problem. Recently, a method called Wirtinger flow (WF) (Candes et al., 2015b) was shown to work remarkably well by using a spectral method for initialization and gradient descent for refinement. It only requires $O(n \log n)$ measurements to recover the signal within $O(mn^2 \cdot \log(1/\epsilon))$ flops. The follow-up work, called Truncated Wirtinger Flow (TWF) (Chen and Candes, 2015), introduced a truncation mechanism to select a subset of samples, which reduces the sample complexity to $O(n)$ and computational complexity to $O(mn \cdot \log(1/\epsilon))$. More recently, Truncated Amplitude Flow (TAF) (Wang et al., 2016b) and Reshaped Wirtinger Flow (RWF) (Zhang and Liang, 2016) used the magnitude of $\langle \mathbf{a}_i, \mathbf{x}^* \rangle$ instead of its square as the observations. Zhang and Liang (2016) proved that RWF enjoys the same sample complexity as TWF even without truncation in gradient steps. TAF (Wang et al., 2016b) used an orthogonality-promoting initialization method, which returns better initial solutions compared with the spectral counterparts. Many stochastic/incremental algorithms such as Incremental Truncated Wirtinger Flow (ITWF) (Kolte and Özgür, 2016), Incremental Reshaped Wirtinger Flow (IRWF) (Zhang et al., 2016b), Stochastic Truncated Amplitude Flow (STAF) (Wang et al., 2016a) have also been developed. However, these stochastic/incremental algorithms fail to improve the computational complexity due to the higher requirements on the step size. While the aforementioned variants of WF do improve the robustness of the original WF algorithm by carefully selecting a subset of samples, they still fail in arbitrary corruption setting.

In a seminal work (Zhang et al., 2016a) for phase retrieval with arbitrary corruption, a Median-TWF algorithm was introduced to utilize the properties of the median estimator to enhance the robustness in such a corruption setting. Nevertheless, it requires $O(n \log n)$ samples in the gradient descent stage in order to recover the unknown signal. The reason is that by using the median truncation mechanism, many uncorrupted samples with large magnitude are discarded and thus it cannot fully utilize the given information to recover both the signal and the corruption. As a result, their method leads to a suboptimal sample complexity. Hand (2017) showed that PhaseLift is robust to corruption, however, its computational complexity is at least cubic in $n$, which is time consuming.

Another line of research that is very related to our work is low-rank matrix/tensor estimation under corruption, that includes robust principal component analysis (Candès et al., 2011; Chandrasekaran et al., 2011; Hsu et al., 2011; Chen et al., 2013; Netrapalli et al., 2014; Gu et al., 2016; Yi et al., 2016), robust low-rank matrix completion (Agarwal et al., 2012; Goldfarb and Qin, 2014; Klopp et al., 2014; Cherapanamjeri et al., 2016), robust tensor decomposition (Gu et al., 2014; Anandkumar et al., 2015). The key idea of these methods is to estimate the unknown low-rank matrix/tensor and the sparse corruption matrix/tensor simultaneously. At a high level, our work is inspired by this line of research.

## 3 Problem Setup

We consider the phase retrieval problem where the observations are contaminated by sparse corruption with arbitrarily large magnitudes and random noises. For concreteness, our analysis will focus on the real-valued Gaussian model. It is worth noting that our proposed algorithm can be directly extended to complex Gaussian model. More specifically, suppose the observations are generated from the following measurement model:

$$y_i = |\mathbf{a}_i^\top \mathbf{x}^*| + \eta_i^* + \epsilon_i, \ 1 \leq i \leq m, \tag{3.1}$$

where $\{\mathbf{a}_i\}_{i=1}^m$ are Gaussian random measurement vectors, each of which is independently drawn from a multivariate normal distribution $N(\mathbf{0}, \mathbf{I}_{n \times n})$, $\mathbf{x}^* \in \mathbb{R}^n$ is the unknown true signal to be recovered, and $\boldsymbol{\eta}^* \in \mathbb{R}^m$ is the arbitrarily large corruption vector that contains at most $\alpha m$ non-zero entries. Through the rest of this paper, we refer to $\alpha$ as the corruption fraction parameter of $\boldsymbol{\eta}^*$. And $\boldsymbol{\epsilon} \in \mathbb{R}^m$ is the noise vector with zero mean and each element $\epsilon_i$ is bounded by $|\epsilon_i| \leq \delta \|\mathbf{x}^*\|_2$ for some constant $\delta$. Similar boundedness assumption on $\epsilon_i$ has been widely made in the literature (Chen and Candes, 2015; Zhang et al., 2016a; Zhang and Liang, 2016; Wang et al., 2016b). When there is no random noise, Model (3.1) will reduce to

$$y_i = |\mathbf{a}_i^\top \mathbf{x}^*| + \eta_i^*, \ 1 \leq i \leq m. \tag{3.2}$$

Note that the same models have been investigated in Zhang et al. (2016a); Hand and Voroninski (2016). Next, following the convention of phase retrieval (Candes et al., 2015b; Chen and Candes, 2015; Zhang and Liang, 2016; Wang et al., 2016b), we define the distance between an estimate $\mathbf{x}$ to the unknown signal $\mathbf{x}^*$ to be:

$$\text{dist}(\mathbf{x}, \mathbf{x}^*) := \min_{\phi \in [0, 2\pi)} \|e^{-j\phi} \mathbf{x} - \mathbf{x}^*\|_2.$$

For real valued data, it can be reduced to $\text{dist}(\mathbf{x}, \mathbf{x}^*) := \min\{\|\mathbf{x} + \mathbf{x}^*\|_2, \|\mathbf{x} - \mathbf{x}^*\|_2\}$.

## 4 The Proposed Algorithm

In this section, we introduce our proposed Robust Wirtinger Flow algorithm in Algorithm 1. Our method is composed of two stages: the initialization stage and the gradient descent stage.

The hard thresholding operator $\mathcal{H}$ in Algorithm 1 is defined as follows:

$$[\mathcal{H}_s(\mathbf{w})]_i := \begin{cases} w_i, & \text{if } |w_i| \geq |\mathbf{w}^{(s)}|, \\ 0, & \text{otherwise}, \end{cases} \tag{4.1}$$

where $\mathbf{w}^{(s)}$ denotes the element whose magnitude is the $s$-largest in $\mathbf{w} \in \mathbb{R}^m$. Note that the thresholding parameter $\widetilde{\alpha}$ is a tuning parameter in real world problems.

**Algorithm 1** Robust Wirtinger Flow
1: **Input:** Observation vector $\mathbf{y} = \{y_i\}_{i=1}^m$, Measurement vectors $\{\mathbf{a}_i\}_{i=1}^m$, Thresholding parameter $\widetilde{\alpha}$, Stepsize $\mu$.

   *Stage I: Initialization*
2: $\boldsymbol{\eta}^{(0)} = \mathcal{H}_{\widetilde{\alpha}m}(\mathbf{y})$, $\widehat{\mathbf{y}} = \mathbf{y} - \boldsymbol{\eta}^{(0)}$
3: $\lambda_0 = \sqrt{\frac{1}{m}\sum_{i=1}^m \widehat{y_i}^2}$
4: $\mathbf{x}^{(0)} = \lambda_0 \widetilde{\mathbf{x}}$ where $\widetilde{\mathbf{x}}$ is the leading eigenvector of
   $$Y := \frac{1}{m}\sum_{i=1}^m \widehat{y_i}^2 \mathbf{a}_i \mathbf{a}_i^\top$$

   *Stage II: Gradient Descent*
5: **for** $t = 0$ to $T-1$ **do**
6:   $\boldsymbol{\eta}^{(t+1)} = \mathcal{H}_{\widetilde{\alpha}m}(\mathbf{y} - \mathcal{A}(\mathbf{x}^{(t)}))$
     where $[\mathcal{A}(\mathbf{x}^{(t)})]_i = |\mathbf{a}_i^\top \mathbf{x}^{(t)}|$
7:   Update $\mathbf{x}^{(t+1)}$ by
   $$\mathbf{x}^{(t)} - \frac{\mu}{m}\sum_{i=1}^m \left(|\mathbf{a}_i^\top \mathbf{x}^{(t)}| + \eta_i^{(t+1)} - y_i\right)\mathrm{sgn}(\mathbf{a}_i^\top \mathbf{x}^{(t)})\mathbf{a}_i$$
8: **end for**
9: **Output:** $\mathbf{x}^{(T)}$

For the ease of presentation, throughout the rest of this paper, we denote by $y_i^* = |\mathbf{a}_i^\top \mathbf{x}^*|$ the non-contaminated component (i.e., without the random noise and arbitrary corruption) of the $i$-th measurement. Let $\mathbf{y}^* = [y_1^*, \ldots, y_m^*]^\top$.

**Initialization Stage:** Our initialization procedure consists of two parts: estimating the magnitude of the true signal $\mathbf{x}^*$ by $\lambda_0$ and estimating the phase of $\mathbf{x}^*$ by $\widetilde{\mathbf{x}}$. The intuition here is to find a good estimator $\widehat{\mathbf{y}}$ for the non-contaminated measurements $\mathbf{y}^*$ and show that our magnitude estimator $\lambda_0$ and phase estimator $\widetilde{\mathbf{x}}$ based on $\widehat{\mathbf{y}}$ would be close to their corresponding values based on the unknown $\mathbf{y}^*$. By taking the hard thresholding operator in Step 2 of Algorithm 1, $\boldsymbol{\eta}^{(0)}$ contains the largest $\widetilde{\alpha}m$ elements in $\mathbf{y}$. Hence $\widehat{\mathbf{y}} = \mathbf{y} - \boldsymbol{\eta}^{(0)}$ is the observation vector with largest $\widetilde{\alpha}m$ elements removed. Suppose the magnitude of the corruption is fairly large, it is reasonable to adopt hard thresholding operator to remove the large entries in $\mathbf{y}$ temporarily. Similar ideas have been used in Yi et al. (2016); Netrapalli et al. (2014).

For the magnitude estimation, in the arbitrary corruption setting, we will later prove that, by conducting the hard thresholding operator as shown in Algorithm 1, as long as the corruption fraction $\alpha$ satisfies certain upper bound condition, $\lambda_0 := \sqrt{(1/m)\sum_{i=1}^m \widehat{y_i}^2}$ concentrates around $\|\mathbf{x}^*\|_2$ with high probability.

In terms of estimating the phase of the unknown signal, following the idea of Candes et al. (2015b), a spectral method is used by computing the leading eigenvector of $Y := (1/m)\sum_{i=1}^m \widehat{y_i}^2 \mathbf{a}_i \mathbf{a}_i^\top$, except that we use $\widehat{y_i}$ instead of $y_i$. Similarly, provided that the corruption fraction $\alpha$ is sufficiently small, we can show that $\widetilde{\mathbf{x}}$ is a good estimator for the phase of $\mathbf{x}^*$.

**Gradient Descent Stage:** In this stage, we propose to solve the following minimization problem under sparsity constraint:

$$\min_{\mathbf{x}\in\mathbb{R}^n, \boldsymbol{\eta}\in\mathbb{R}^m} L(\mathbf{x}, \boldsymbol{\eta}) = \frac{1}{2m}\sum_{i=1}^{m}\left(y_i - |\mathbf{a}_i^\top \mathbf{x}| - \eta_i\right)^2,$$

$$\text{subject to } \|\boldsymbol{\eta}\|_0 \leq \widetilde{\alpha} m.$$

It is a natural idea to use a gradient descent algorithm to solve the above problem. Due to the sparsity constraint on $\boldsymbol{\eta}$, we need to apply the hard thresholding operator in (4.1) again to get our new corruption estimator $\boldsymbol{\eta}^{(t+1)}$. According to our model in (3.1) we have $y_i = |\mathbf{a}_i^\top \mathbf{x}^*| + \eta_i^* + \epsilon_i$, it is reasonable to use the following update for $\boldsymbol{\eta}^{(t+1)}$:

$$\boldsymbol{\eta}^{(t+1)} = \mathcal{H}_{\widetilde{\alpha} m}\big(\mathbf{y} - \mathcal{A}(\mathbf{x}^{(t)})\big),$$

where $\big[\mathcal{A}(\mathbf{x}^{(t)})\big]_i = |\mathbf{a}_i^\top \mathbf{x}^{(t)}|$. For updating $\mathbf{x}^{(t+1)}$, one can compute the gradient of $L(\mathbf{x}, \boldsymbol{\eta})$ with respect to $\mathbf{x}$ as follows:

$$\nabla_{\mathbf{x}} L(\mathbf{x}, \boldsymbol{\eta}) = \tfrac{1}{m}\sum_{i=1}^{m}(|\mathbf{a}_i^\top \mathbf{x}| + \eta_i - y_i)\cdot \operatorname{sgn}(\mathbf{a}_i^\top \mathbf{x})\cdot \mathbf{a}_i,$$

which naturally leads to a gradient update for $\mathbf{x}^{(t+1)}$ as:

$$\mathbf{x}^{(t)} - \tfrac{\mu}{m}\sum_{i=1}^{m}\big(|\mathbf{a}_i^\top \mathbf{x}^{(t)}| + \eta_i^{(t+1)} - y_i\big)\cdot \operatorname{sgn}\big(\mathbf{a}_i^\top \mathbf{x}^{(t)}\big)\cdot \mathbf{a}_i,$$

where $\mu$ is the step size. Note that for our model in (3.1), i.e., $y_i = |\mathbf{a}_i^\top \mathbf{x}^*| + \eta_i^* + \epsilon_i$, when $\mathbf{x}^{(t)}$ gets closer to $\mathbf{x}^*$, our estimation for $\boldsymbol{\eta}^*$ also becomes more accurate. Thus by taking the gradient update for only $\mathbf{x}$, we manage to get precise estimation results for both $\mathbf{x}^*$ and $\boldsymbol{\eta}^*$.

## 5 Main Theory

In this section, we provide the main theory about Algorithm 1, including the linear rate convergence analysis and sharper statistical results for robust phase retrieval with arbitrary corruption.

**Theorem 5.1.** Consider the phase retrieval problem with both arbitrary corruption and bounded noises defined in (3.1). Let $C_0, C_1, C_2, C_3$ be some universal constants. For any signal $\mathbf{x}^* \in \mathbb{R}^n$, Gaussian measurement vectors $\{\mathbf{a}_i\}_{i=1}^m$ drawn from i.i.d $N(\mathbf{0}, \mathbf{I}_{n\times n})$, let $\boldsymbol{\epsilon} \in \mathbb{R}^m$ be a bounded noise vector with $|\epsilon_i| \leq \delta \cdot \|\mathbf{x}^*\|_2$, and $\boldsymbol{\eta}^*$ be the $\alpha m$-sparse corruption vector with arbitrary large magnitudes, suppose the sparse parameter $\alpha$ satisfies $\alpha \leq C_0/\log m$, if $m \geq C_1 \cdot n$, then $\mathbf{x}^{(0)}$ generated by Stage I of Algorithm 1 satisfies

$$\operatorname{dist}(\mathbf{x}^{(0)}, \mathbf{x}^*) \leq \frac{1}{10}\|\mathbf{x}^*\|_2,$$

and the output $\mathbf{x}^{(t)}$ from Stage II of Algorithm 1 satisfies

$$\operatorname{dist}(\mathbf{x}^{(t)}, \mathbf{x}^*) \leq \underbrace{\frac{1}{10}\Big(1 - \frac{\mu}{2}\Big)^t \cdot \|\mathbf{x}^*\|_2}_{\text{optimization error}} + \underbrace{C_4 \mu \cdot \|\boldsymbol{\epsilon}\|_\infty}_{\text{statistical error}},$$

with probability at least $1 - C_2 e^{-C_3 m} - 4/m$.

**Remark 5.2.** Theorem 5.1 provides theoretical guarantees on the performance of our algorithm for robust phase retrieval with arbitrary corruption and random noise. As we can see, the distance between $\mathbf{x}^{(t)}$ output by our algorithm and $\mathbf{x}^*$ is upper bounded by two terms: the optimization error and the statistical error. Due to the linear convergence rate of our proposed algorithm, at most $O(\log(1/\epsilon))$ iterations are needed in order to achieve $\epsilon$-error in optimization. And the overall computational complexity for our algorithm is $O(mn\log(1/\epsilon))$, which matches the computational cost of state-of-the-art methods (Chen and Candes, 2015; Wang et al., 2016b; Zhang and Liang, 2016). In terms of statistical error, Algorithm 1 also matches the state-of-the-art statistical rate in the order of $O(\|\boldsymbol{\epsilon}\|_\infty)$ or $O(\|\boldsymbol{\epsilon}\|_2/\sqrt{m})$ (Chen and Candes, 2015; Zhang and Liang, 2016; Wang et al., 2016b; Zhang et al., 2016a).

**Remark 5.3.** Theorem 5.1 suggests that Algorithm 1 achieves an optimal sample complexity of $O(n)$ in both initialization stage and gradient descent stage. This matches the best known results of those non-robust phase retrieval algorithms such as TWF (Chen and Candes, 2015), TAF (Wang et al., 2016b) and RWF (Zhang and Liang, 2016). In contrast, Median-TWF method (Zhang et al., 2016a) only achieves a sample complexity of $O(n\log n)$. In terms of corruption tolerance, we notice that Median-TWF (Zhang et al., 2016a), is robust to constant corruption fraction, while our proposed algorithm requires that the corruption fraction $\alpha$ satisfies $\alpha \leq C_0/\log m$. We conjecture that such a logarithmic factor is just an artifact of our proof. In the experiments, we observe that our algorithm is as robust as Median-TWF. We leave the removal of this logarithmic factor as a future work.

An immediate consequence of Theorem 5.1 is the following corollary, which suggests that our algorithm can exactly recover the unknown signal $\mathbf{x}^*$ with arbitrary large corruption in the noise-free model.

**Corollary 5.4.** Consider the phase retrieval problem with arbitrary corruption defined in (3.2). Under the same condition as in Theorem 5.1, $\mathbf{x}^{(0)}$ generated by Stage I of Algorithm 1 satisfies

$$\text{dist}(\mathbf{x}^{(0)}, \mathbf{x}^*) \leq \frac{1}{10}\|\mathbf{x}^*\|_2,$$

and the output $\mathbf{x}^{(t)}$ from Stage II of Algorithm 1 satisfies

$$\text{dist}(\mathbf{x}^{(t)}, \mathbf{x}^*) \leq \frac{1}{10}\left(1 - \frac{\mu}{2}\right)^t \cdot \|\mathbf{x}^*\|_2,$$

with probability at least $1 - C_4 e^{-C_3 m} - 4/m$.

## 6 Proof Sketch of the Main Theorem

In this section, we highlight the proof sketch of the main theory in Section 5. The detailed proofs can be found in the Appendix.

### 6.1 Robust Initialization

In this subsection we briefly demonstrate how our initialization procedure showed in Stage I of Algorithm 1 will generate an initial solution close enough to the true signal.

First we investigate the composition of $\widehat{\mathbf{y}}$. Since $\boldsymbol{\eta}^{(0)} = \mathcal{H}_{\widetilde{\alpha} m}(\mathbf{y})$, where $\mathcal{H}$ denotes for the hard thresholding operator defined in (4.1) and $\widehat{\mathbf{y}} = \mathbf{y} - \boldsymbol{\eta}^{(0)}$ by our algorithm procedure. It can be shown that the following two claims hold:

$$\text{Claim 1: } \widehat{\mathbf{y}} - \mathbf{y}^* - \boldsymbol{\epsilon} = \boldsymbol{\eta}^* - \boldsymbol{\eta}^{(0)},$$
$$\text{Claim 2: } \|\widehat{\mathbf{y}} - \mathbf{y}^* - \boldsymbol{\epsilon}\|_\infty \leq 2\|\mathbf{y}^* + \boldsymbol{\epsilon}\|_\infty.$$

Given those two claims, we have the following lemma to ensure the estimation accuracy of the magnitude of signal, i.e., $\|\mathbf{x}^*\|_2$.

**Lemma 6.1.** Let $\boldsymbol{\eta}^{(0)}, \widehat{\mathbf{y}}$ be defined as in Stage I of Algorithm 1, suppose $\boldsymbol{\eta}^*$ is $\alpha m$-sparse, $\boldsymbol{\eta}^{(0)}$ is $\widetilde{\alpha} m$-sparse and the corruption sparse parameter $\alpha$ satisfies $\alpha \leq c / \log m$, and the additional noise is bounded as $|\epsilon_i| \leq \delta \|\mathbf{x}^*\|_2$, for the model defined in (3.1) we have

$$(0.99 - 5\delta)\|\mathbf{x}^*\|_2^2 \leq \frac{1}{m}\sum_{i=1}^{m} \widehat{y}_i^2 \leq (1.02 + 6\delta)\|\mathbf{x}^*\|_2^2,$$

with probability at least $1 - 4e^{-c_1 m} - 2/m$ where $\delta$ is a problem dependent constant.

Next we evaluate the accuracy for estimating the phase of $\mathbf{x}^*$. In Algorithm 1 we have $\mathbf{Y} := \frac{1}{m}\sum_{i=1}^{m} \widehat{y}_i^2 \mathbf{a}_i \mathbf{a}_i^\top$. Without loss of generality, we assume $\|\mathbf{x}^*\|_2 = 1$. Since $\widehat{\mathbf{y}}$ exclude those observations which are hard-thresholded, we can rewrite each $\widehat{y}_i$ as

$$\widehat{y}_i = y_i \cdot \mathbb{1}\{y_i \leq \theta_{1-\gamma\alpha}(\{y_i\})\},$$

where $\theta_p(\cdot)$ represents the $p$-quantile of the given samples. By utilizing the quantile-based concentration inequality and matrix perturbation theory (the Davis-Kahan $\sin\Theta$ Theorem (Davis, 1963; Yu et al., 2015)), we have the following lemma, which characterizes the difference between $\widetilde{\mathbf{x}}$, the leading eigenvector of $\mathbf{Y}$, and $\mathbf{x}^*$.

**Lemma 6.2.** Consider the model defined in (3.1), suppose $\boldsymbol{\eta}^*$ is $\alpha m$-sparse, the additional noise is bounded as $|\epsilon_i| \leq \delta \|\mathbf{x}^*\|_2$ and the sparse parameter $\alpha$ is smaller than an universal constant $c$. If $m \geq c_1 \cdot n$, the leading eigenvector of $\mathbf{Y}$, denoted by $\widetilde{\mathbf{x}}$, satisfies

$$\text{dist}(\widetilde{\mathbf{x}}, \mathbf{x}^*) \leq \widetilde{\tau},$$

with probability at least $1 - c_2 e^{-c_3 m}$, where $\widetilde{\tau}$ is a given constant.

Finally consider the distance between $\mathbf{x}^{(0)}$ and $\mathbf{x}^*$ when $\|\mathbf{x}^*\|_2 \neq 1$,

$$\begin{aligned} \text{dist}(\mathbf{x}^{(0)}, \mathbf{x}^*) &\leq \text{dist}(\lambda_0 \widetilde{\mathbf{x}}, \|\mathbf{x}^*\|_2 \cdot \widetilde{\mathbf{x}}) + \text{dist}(\|\mathbf{x}^*\|_2 \cdot \widetilde{\mathbf{x}}, \mathbf{x}^*) \\ &\leq \max\{\sqrt{1.02 + 6\delta} - 1, 1 - \sqrt{0.99 - 5\delta}\}\|\mathbf{x}^*\|_2 + \widetilde{\tau}\|\mathbf{x}^*\|_2 \\ &\leq \frac{1}{10}\|\mathbf{x}^*\|_2, \end{aligned}$$

where the last inequality can be easily achieved by choosing $\delta, \widetilde{\tau}$ to be sufficiently small. This completes the proof.

## 6.2 Convergence Analysis

In this subsection, we show that once the initial solution falls in the neighborhood of $\mathbf{x}^*$, the Stage II of our algorithm guarantees a linear convergence towards the true signal $\mathbf{x}^*$. Denote the support for $\boldsymbol{\eta}^*$ as $\Omega^*$ and the support for $\boldsymbol{\eta}^{(t+1)}$ as $\Omega'$. By discussing the support each sample index $i$ falls in, we decompose the gradient $\nabla_{\mathbf{x}} L(\mathbf{x}^{(t)}, \boldsymbol{\eta}^{(t+1)})$ into two parts:

**Approximate Gradient**:

$$\mathbf{g}(\mathbf{x}^{(t)}, \boldsymbol{\eta}^{(t+1)}) = \frac{1}{m} \sum_{i \notin \Omega'} \left( |\mathbf{a}_i^\top \mathbf{x}^{(t)}| - |\mathbf{a}_i^\top \mathbf{x}^*| \right) \operatorname{sgn}(\mathbf{a}_i^\top \mathbf{x}^{(t)}) \mathbf{a}_i.$$

**Residual Gradient**:

$$\nabla_{\mathbf{x}} L(\mathbf{x}^{(t)}, \boldsymbol{\eta}^{(t+1)}) - \mathbf{g}(\mathbf{x}^{(t)}, \boldsymbol{\eta}^{(t+1)}) = -\frac{1}{m} \sum_{i \in \Omega^* \setminus \Omega'} \eta_i^* \cdot \operatorname{sgn}(\mathbf{a}_i^\top \mathbf{x}^{(t)}) \mathbf{a}_i - \frac{1}{m} \sum_{i \notin \Omega'} \epsilon_i \cdot \operatorname{sgn}(\mathbf{a}_i^\top \mathbf{x}^{(t)}) \mathbf{a}_i.$$

By the gradient descent update rules we can easily have

$$\operatorname{dist}(\mathbf{x}^{(t+1)}, \mathbf{x}^*) \leq \underbrace{\left\| \mathbf{h} - \mu \cdot \mathbf{g}(\mathbf{x}^{(t)}, \boldsymbol{\eta}^{(t+1)}) \right\|_2}_{\text{approximate distance}} + \mu \underbrace{\left\| \nabla_{\mathbf{x}} L(\mathbf{x}^{(t)}, \boldsymbol{\eta}^{(t+1)}) - \mathbf{g}(\mathbf{x}^{(t)}, \boldsymbol{\eta}^{(t+1)}) \right\|_2}_{\text{residual gradient error}}. \tag{6.1}$$

For the approximate distance term we have the following lemma:

**Lemma 6.3.** For the approximate distance defined in (6.1), if $m \geq c_0 n$ and the corruption sparsity parameter $\alpha$ satisfies $\alpha \leq c/\log m$ then with probability at least $1 - 4e^{-c_1 m} - 1/m$, we have

$$\left\| \mathbf{h} - \mu \cdot \mathbf{g}(\mathbf{x}^{(t)}, \boldsymbol{\eta}^{(t+1)}) \right\|_2^2 \leq \left( 1 - 2\mu(0.73 - 3\delta) + \mu^2(1+\delta)^2 \right) \|\mathbf{h}\|_2^2,$$

holds for all $\mathbf{h} \in \mathbb{R}^n$ satisfying $\|\mathbf{h}\|_2 \leq \|\mathbf{x}^*\|_2/10$.

As for the two terms in the residual gradient error, we have the following two lemmas characterizing their upper bound respectively:

**Lemma 6.4.** Denote the support for $\boldsymbol{\eta}^*$ as $\Omega^*$ and the support for $\boldsymbol{\eta}^{(t+1)}$ as $\Omega'$, suppose $\mathbf{a}_i \in \mathbb{R}^n, i = 1, \ldots, m$ are Gaussian vectors independently drawn from $N(\mathbf{0}, \mathbf{I}_{n \times n})$, if the corruption sparsity parameter $\alpha$ satisfies $\alpha \leq c/\log m$, then with probability at least $1 - 2e^{-c_3 m} - 1/m$, the following

$$\left\| \frac{1}{m} \sum_{i \in \Omega^* \setminus \Omega'} \eta_i^* \cdot \operatorname{sgn}(\mathbf{a}_i^\top \mathbf{x}^{(t)}) \mathbf{a}_i \right\|_2 \leq 0.02 \cdot \|\mathbf{h}\|_2 + 2\sqrt{\alpha(1+\delta)} \cdot \|\boldsymbol{\epsilon}\|_\infty,$$

holds for all non-zero vectors $\mathbf{h} \in \mathbb{R}^n$.

**Lemma 6.5.** Denote the support for $\boldsymbol{\eta}^{(t+1)}$ as $\Omega'$, suppose $\mathbf{a}_i \in \mathbb{R}^n, i = 1, \ldots, m$ are Gaussian vectors independently drawn from $N(\mathbf{0}, \mathbf{I}_{n \times n})$, we have

$$\left\| \frac{1}{m} \sum_{i \notin \Omega'} \epsilon_i \cdot \operatorname{sgn}(\mathbf{a}_i^\top \mathbf{x}^{(t)}) \mathbf{a}_i \right\|_2 \leq \sqrt{1+\delta} \cdot \|\boldsymbol{\epsilon}\|_\infty,$$

with probability at least $1 - 2e^{-c_4 m}$.

Combining Lemmas 6.3, 6.4 and 6.5, we obtain the following inequality describing the contraction between consecutive iterations:

$$\text{dist}(\mathbf{x}^{(t+1)}, \mathbf{x}^*) \leq (1 - \mu/2) \cdot \|\mathbf{h}\|_2 + c_4 \mu \cdot \|\boldsymbol{\epsilon}\|_\infty,$$

where $\mu \leq \mu_0$ with $\mu_0$ as a positive universal constant, $\delta, \alpha$ is chosen to be sufficiently small. The final conclusion in Theorem 5.1 is obtained by iteratively conducting the above contraction formula with a mathematical induction argument.

## 7 Experiments

In this section, we evaluate the performance of our Robust-WF algorithm against other state-of-the-art baseline algorithms. We conduct our experiments on both synthetic and real data set.

### 7.1 Baseline Methods

We compare our algorithm with the following baseline methods: **RWF**: Reshaped Wirtinger Flow (Zhang and Liang, 2016), **TAF**: Truncated Amplitude Flow (Wang et al., 2016b), **TWF**: Truncated Wirtinger Flow (Chen and Candes, 2015), and **Median-TWF**: Median Truncated Wirtinger Flow (Zhang et al., 2016a).

Note that here we did not compare with the original Wirginter Flow algorithm (Candes et al., 2015b) since both its initialization stage and gradient descent stage do not involve any truncation or robust estimation technique, thus does not work well when corruption exists. Also we did not compare with Candes et al. (2013) since it is a convex method with much higher computational complexity.

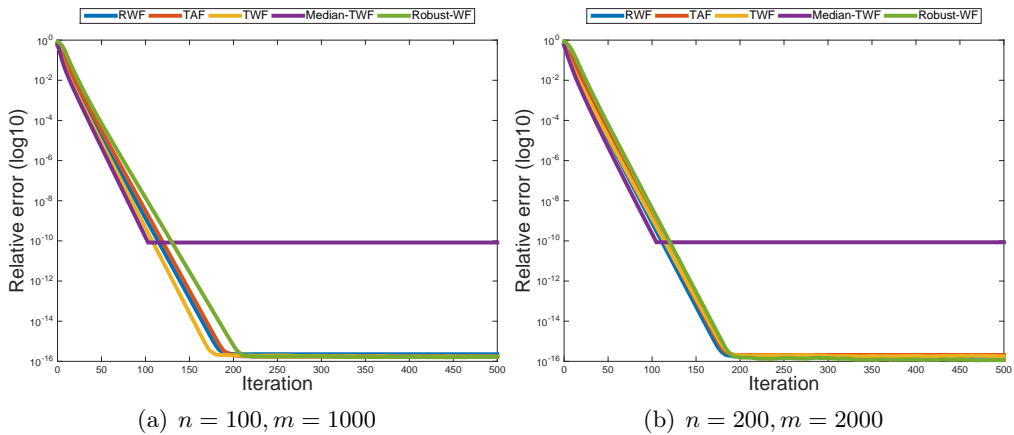

**Figure 1:** Relative error with respect to the iteration count for all baseline algorithms and our proposed algorithm under no corruption noise-free circumstances.

### 7.2 Parameter Settings

For all algorithms, we run a fixed number of iterations $T = 250$ and the number of power iterations in initialization stage is also fixed to 200. We set $\widetilde{\alpha} = 2\alpha$ for our Robust-WF algorithm and all the truncation parameters in those baseline algorithms to the suggested values in respective papers. The step size is tuned for each algorithm.

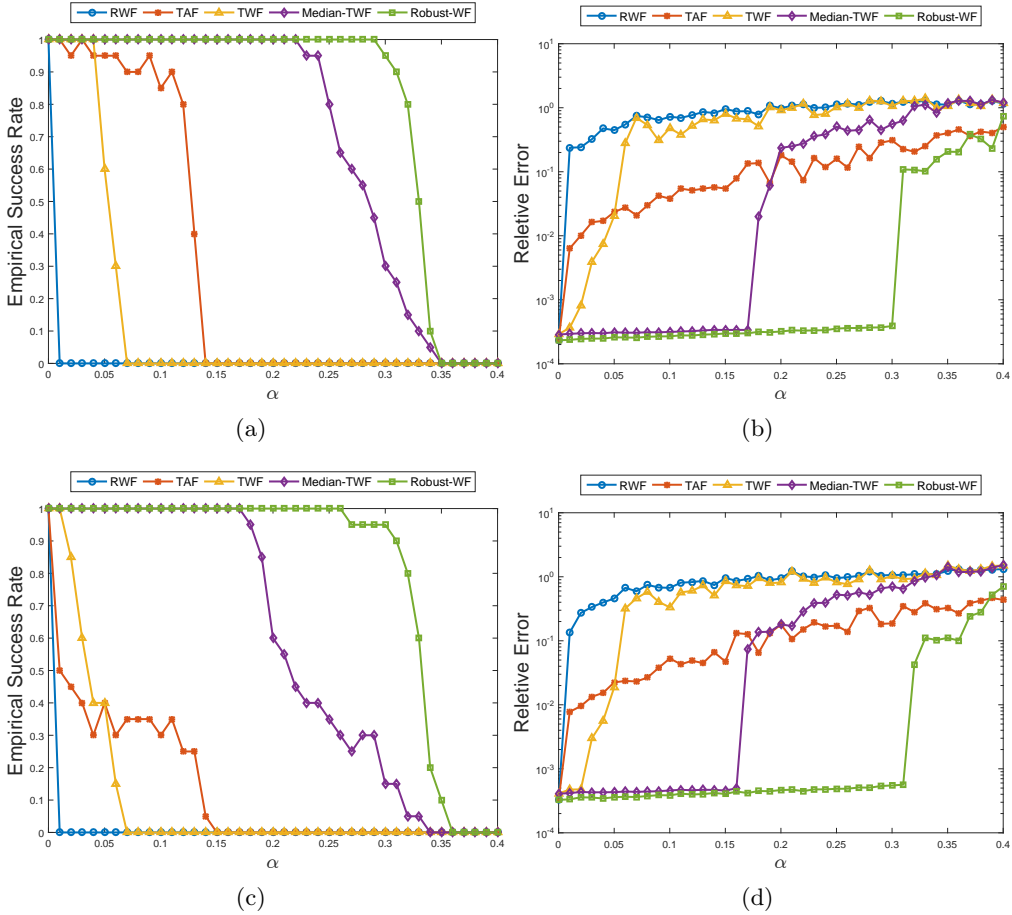

**Figure 2:** (a) Empirical success rate of exact recovery against the corruption fraction $\alpha$ for all algorithms where $n = 200, m = 2000$ under 20 replications; (b) Relative error against the corruption fraction $\alpha$ for all algorithms where $n = 200, m = 2000$ under 20 replications. (c) Empirical success rate of exact recovery against the corruption fraction $\alpha$ for all algorithms where $n = 100, m = 1000$ under 20 replications; (d) Relative error against the corruption fraction $\alpha$ for all algorithms where $n = 100, m = 1000$ under 20 replications.

## 7.3 Synthetic Data

In each setting, we generate a Gaussian measurement matrix $\mathbf{A} \in \mathbb{R}^{m \times n}$ with rows drawn independently from a standard multivariate normal distribution $N(\mathbf{0}, \mathbf{I})$. The true signal $\mathbf{x}^*$ is also generated from an independent standard multivariate normal distribution $N(\mathbf{0}, \mathbf{I})$. We generate a sparse corruption vector $\boldsymbol{\eta}^*$ with at most $\alpha m$ non-zero entries. Each of the non-zero entry comes with a random corruption in the magnitude of $0.5\|\mathbf{x}^*\|_2$. The random noise $\boldsymbol{\epsilon}$ is generated independently from a uniform distribution $U(0, p)$ with $p$ ranging from 0 to 2.

We evaluate the performance of the algorithms based on relative errors, i.e., $\mathrm{dist}(\mathbf{x}^{(T)}, \mathbf{x}^*)/\|\mathbf{x}^*\|_2$. In noise-free setting, we also consider the empirical success rate, which is defined as the ratio of

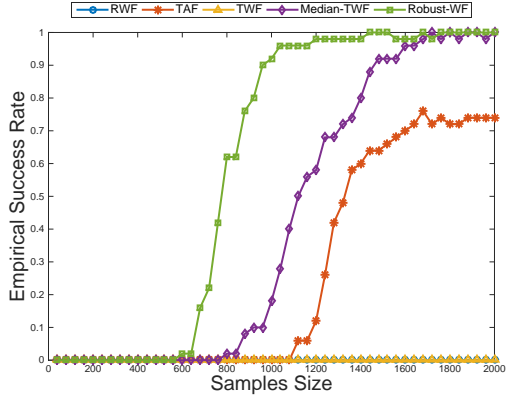

**Figure 3:** Empirical success rate of exact recovery against the sample size $m$ for all algorithms where $n = 200, \alpha = 0.05$ under 20 replications.

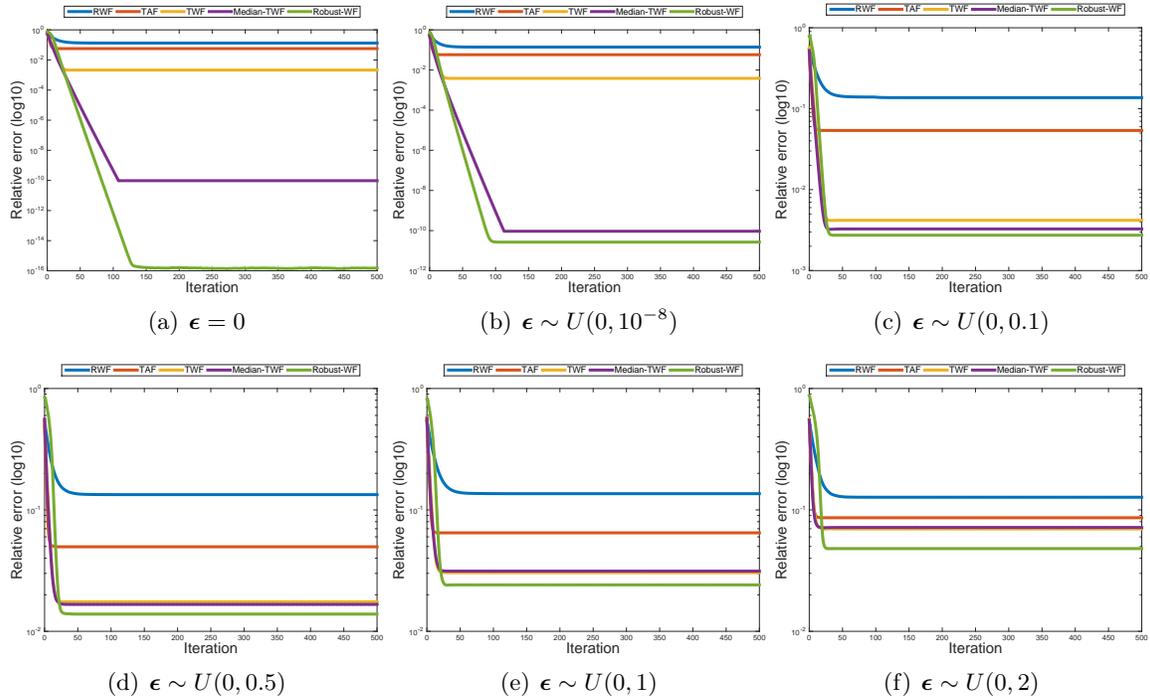

**Figure 4:** Convergence plot (relative error against iterations) for all algorithms with respect to different level of noise in settings where $n = 200, m = 2000, \alpha = 0.05, \|\boldsymbol{\eta}^*\|_\infty \leq 0.2 \cdot \|\mathbf{x}^*\|_2$.

successful trials against the total number of trails. Specifically, a trial is declared successful if the output $\mathbf{x}^{(T)}$ satisfies $\text{dist}(\mathbf{x}^{(T)}, \mathbf{x}^*) \leq 10^{-8}$. We test the performance of all algorithms by ranging $\alpha$ from 0 to 0.4 with an increment of 0.01 in two settings: $n = 100, m = 1000$ and $n = 200, m = 2000$.

Figure 1 shows the case where no corruption and no noise are involved with $n = 100, m = 1000$. All baseline algorithms and our proposed algorithm achieve similar performance except for Median-TWF, which is less accurate compared with other approaches in this case.

Figure 2 (a) and (c) illustrate the performances of all algorithms in the noise-free setting. We

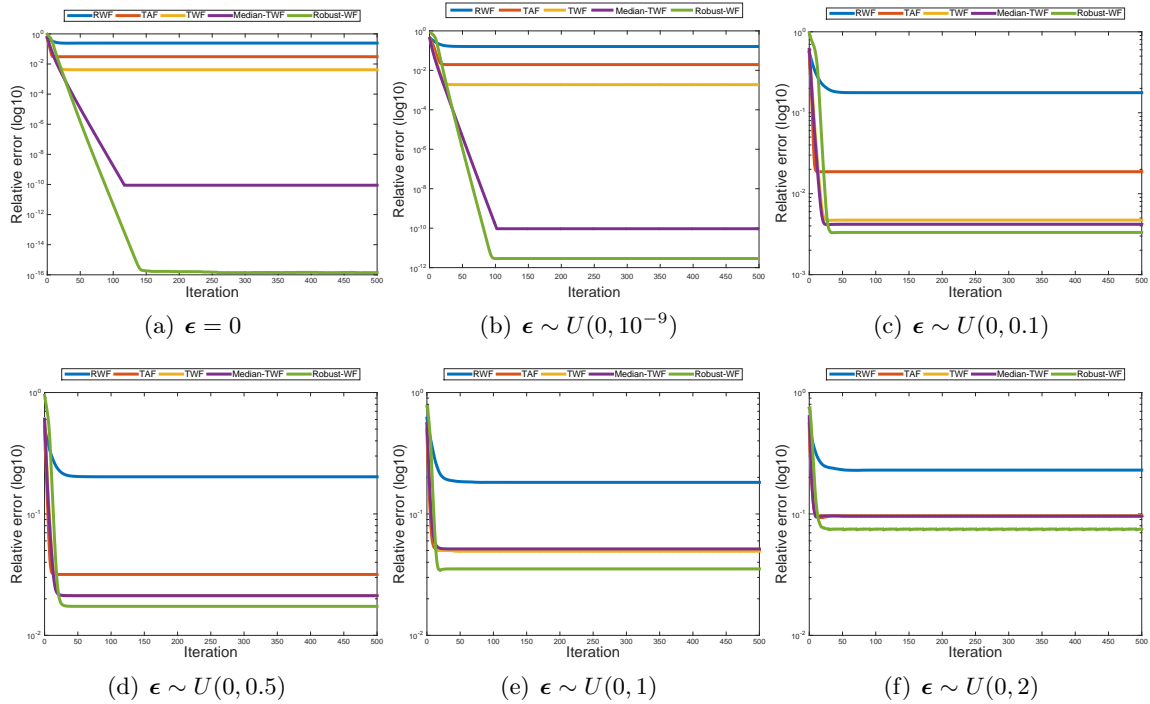

**Figure 5:** Convergence plots (relative error against iterations) for all algorithms with respect to different level of noise in settings where $n = 100, m = 1000, \alpha = 0.05, \|\boldsymbol{\eta}^*\|_\infty \leq 0.2 \cdot \|\mathbf{x}^*\|_2$.

**Table 1:** Summary of the Test Images.

| Images | Dimensions | #Modulations |
|---|---|---|
| Lenna | $512 \times 512 \times 3$ | 12 |
| Stanford | $1280 \times 320 \times 3$ | 12 |
| Galaxy | $1920 \times 1080 \times 3$ | 12 |

* The last coordinate 3 in dimensions refers to the three color bands (R/G/B) of the images.

**Table 2:** Relative Error on the Recovery of Three Testing Images.

| Methods | Lenna | Stanford | Galaxy |
|---|---|---|---|
| RWF | $9.49 \times 10^{-2}$ | $6.54 \times 10^{-2}$ | $8.37 \times 10^{-2}$ |
| TAF | $1.10 \times 10^{-3}$ | $1.50 \times 10^{-3}$ | $1.19 \times 10^{-3}$ |
| TWF | $7.20 \times 10^{-3}$ | $4.70 \times 10^{-3}$ | $6.20 \times 10^{-3}$ |
| Median-TWF | $4.32 \times 10^{-8}$ | $3.31 \times 10^{-8}$ | $2.90 \times 10^{-10}$ |
| Robust-WF | $\mathbf{1.79 \times 10^{-8}}$ | $\mathbf{7.83 \times 10^{-13}}$ | $\mathbf{2.75 \times 10^{-12}}$ |

* Bold numbers represent the best result.

plot the empirical success rate of all five algorithms including ours against the corruption fraction under 20 replications. As we can see from the figure, our proposed algorithm actually out-performs all the other baseline methods. Non-robust algorithms like TWF, RWF and TAF starts to fail quite early even when the corruption fraction is quite small. Median-TWF algorithm also fails when $\alpha$ goes to around 0.23, while our proposed Robust-WF can still successfully recover the signal when $\alpha$ is around 0.3. Figure 2 (b) and (d) illustrate the performances of all algorithms in the noisy setting. We plot the relative error of all algorithms against the corruption fraction under 20 replications. Again, it is obvious from the figure that our proposed algorithm achieves the best relative error compared with other state-of-the-art algorithms.

Figure 3 demonstrates the sample complexity performances of all algorithms in the noise-free setting. We can see that our proposed algorithm achieves better sample complexity over Median-TWF as suggested in our theory. For simplicity, we omit the result for the other setting which is similar.

In Figures 4 and 5, we further investigate the convergence result of our algorithm and the baseline methods. We can see that our proposed algorithm indeed enjoys a linear rate of convergence. This is consistent with our theoretical result in Theorem 5.1. Moreover, our proposed method outperforms all the other baseline algorithms in different noise settings. It can be seen that when the magnitude of the uniformly distributed noise varies from 0 to 2, our algorithm always achieves smallest recovery error over other baselines. Particularly, in noise-free case, even though both our algorithm and Median-TWF meet the criterion for exact recovery (i.e., $\text{dist}(\mathbf{x}^{(T)}, \mathbf{x}^*) \leq 10^{-8}$), the error achieved by our Robust-WF algorithm is significantly smaller than Median-TWF algorithm.

## 7.4 Real Data

We also evaluate the performance of our algorithm on the recovery of real images from the Fourier intensity measurements (two dimensional Coded Diffraction Patterns model). The Coded Diffraction Patterns (CDP) model is a type of physically realizable measurements of images with random masks (see details in Candes et al. (2015a,b); Chen and Candes (2015)). Suppose $\mathbf{x}^* \in \mathbb{R}^n$ is the vectorization of a real image matrix (consider only one color band for each $\mathbf{x}^*$), the CDP model basically collects the magnitude of the discrete Fourier transform (DFT) of $K$ modulations of the signal $\mathbf{x}^*$. Specifically,

$$\mathbf{y}^{(k)} = \left|\mathbf{F}\mathbf{D}^{(k)}\mathbf{x}^*\right|, \ 1 \leq k \leq K,$$

where $\mathbf{F}$ stands for the discrete Fourier transform matrix and $\mathbf{D}^{(k)}$ stands for a diagonal phase delay matrix with its diagonal entries uniformly sampled from $\{1, -1, j, -j\}$. Under the above definition, a total number of $m = nK$ observations are collected. In this experiment, we set $K = 12$ and evaluate all the algorithms by three real world benchmark images. Details about the benchmark images can be found in Table 1. We consider the corrupted CDP model where additional random corruption are imposed upon the CDP model to test the recovery performance of different phase retrieval algorithms. In detail, we randomly selected 5% of the total pixels and impose corruption with magnitudes up to the level of $\|\mathbf{x}^*\|_2$. Table 2 demonstrates the performance of our algorithm and other baseline methods on the recovery of the real images. We can see that for all three benchmark images, our proposed Robust-WF algorithm achieves the smallest relative recovery error. Specifically, on the Lenna image, our algorithm and Median-TWF achieve $10^{-8}$ error, $(1.79 \times 10^{-8}$ versus $4.32 \times 10^{-8})$, while on Stanford and Galaxy images, our algorithm achieves much smaller error than Median-TWF by several orders of magnitude ($10^{-13}$ versus $10^{-8}$ on Stanford, and $10^{-12}$ versus $10^{-10}$ on Galaxy). This clearly demonstrates the superiority of our algorithm over other baseline methods.

## 8 Conclusions and Future Work

In this paper, we proposed a Robust-WF algorithm for phase retrieval with arbitrary corruption. Both theoretical analysis and experiments were conducted to show the superiority of our algorithm. In our future work, we aim to improve our analysis to get rid of the logarithmic factor in the corruption tolerance of our algorithm.

## A Proof of Theorem 5.1

We prove our main theorem in this section. For the ease of expression, we define $\gamma := \widetilde{\alpha}/\alpha$ in the following proof.

### A.1 Robust Initialization

Here we prove that our initialization procedure showed in Stage I of Algorithm 1 will generate an initial solution close enough to the true signal as long as the fraction of corruption $\alpha$ is small enough.

Let $\boldsymbol{\eta}^{(0)} = \mathcal{H}_{\gamma\alpha m}(\mathbf{y})$, where $\mathcal{H}$ denotes for the hard thresholding operator defined in (4.1) and $\widehat{\mathbf{y}} = \mathbf{y} - \boldsymbol{\eta}^{(0)}$ as our initial guess for the true non-contaminated observations $\mathbf{y}^*$. First we want to make two claims:

$$\textbf{Claim 1: } \widehat{\mathbf{y}} - \mathbf{y}^* - \boldsymbol{\epsilon} = \boldsymbol{\eta}^* - \boldsymbol{\eta}^{(0)}, \tag{A.1}$$

$$\textbf{Claim 2: } \|\widehat{\mathbf{y}} - \mathbf{y}^* - \boldsymbol{\epsilon}\|_\infty \leq 2\|\mathbf{y}^* + \boldsymbol{\epsilon}\|_\infty. \tag{A.2}$$

Claim 1 can be shown since by model definition (3.1) we have $\widehat{\mathbf{y}} = \mathbf{y} - \boldsymbol{\eta}^{(0)} = \mathbf{y}^* + \boldsymbol{\eta}^* + \boldsymbol{\epsilon} - \boldsymbol{\eta}^{(0)}$. To show Claim 2 we need to discuss the subset that each data sample belongs to. Denote the support for $\boldsymbol{\eta}^*$ as $\Omega^*$ and the support for $\boldsymbol{\eta}^{(0)}$ as $\Omega$ and we have the following three cases:

- $i \in \Omega$ : Since $i$ belongs to the support for $\boldsymbol{\eta}^{(0)}$, we have $\eta_i^{(0)} = y_i$ and thus $\widehat{y}_i - y_i^* - \epsilon_i = -y_i^* - \epsilon_i$.

- $i \notin \Omega, i \in \Omega^*$ : By model definition (3.1) we have $y_i = y_i^* + \eta_i^* + \epsilon_i$. Since $i$ does not belong to the support for $\boldsymbol{\eta}^{(0)}$, $y_i$ is not in the $\gamma\alpha m$-largest magnitude elements in $\mathbf{y}$ in order to be outside of $\Omega$, i.e., $|y_i| \leq |\mathbf{y}^{(\gamma\alpha m)}|$. Therefore we must have $\eta_i^* \leq 2\|\mathbf{y}^*+\boldsymbol{\epsilon}\|_\infty$, otherwise we would have $|y_i| = |y_i^* + \eta_i^* + \epsilon_i| \geq \|\mathbf{y}^* + \boldsymbol{\epsilon}\|_\infty \geq |\mathbf{y}^{(\gamma\alpha m)}|$ for any $\gamma > 1$ which conflicts the previous conclusion. Also $i \notin \Omega$ means $\eta_i^{(0)} = 0, \widehat{y}_i = y_i$ and hence $\widehat{y}_i - y_i^* - \epsilon_i = \eta_i^* \leq 2\|\mathbf{y}^* + \boldsymbol{\epsilon}\|_\infty$.

- $i \notin \Omega, i \notin \Omega^*$ : Since $i$ does not belong to either the support for $\boldsymbol{\eta}^{(0)}$ or the support for $\boldsymbol{\eta}^*$, we immediately have $\eta_i^* = 0, \widehat{y}_i = y_i$. Note that by model definition (3.1) we have $y_i = y_i^* + \epsilon_i$ and hence $\widehat{y}_i - y_i^* - \epsilon_i = 0$.

According to the above discussion we immediately have $\|\widehat{\mathbf{y}} - \mathbf{y}^* - \boldsymbol{\epsilon}\|_\infty \leq 2\|\mathbf{y}^* + \boldsymbol{\epsilon}\|_\infty$. Given the above two claims, now we first estimate the magnitude of $\mathbf{x}^*$ as

$$\lambda_0 = \sqrt{\frac{1}{m}\sum_{i=1}^{m}\widehat{y}_i^2}.$$

By Lemma 6.1, we have

$$(0.99 - 5\delta)\|\mathbf{x}^*\|_2^2 \leq \frac{1}{m}\sum_{i=1}^m \widehat{y}_i^2 \leq (1.02 + 6\delta)\|\mathbf{x}^*\|_2^2,$$

Next we estimate the direction of $\mathbf{x}^*$. Without loss of generality, we assume $\|\mathbf{x}^*\|_2 = 1$. By Lemma 6.2 we have the difference between $\widetilde{\mathbf{x}}$, the leading eigenvector of $\mathbf{Y}$, and $\mathbf{x}^*$ satisfies:

$$\text{dist}(\widetilde{\mathbf{x}}, \mathbf{x}^*) \leq \widetilde{\tau},$$

Finally consider the distance between $\mathbf{x}^{(0)}$ and $\mathbf{x}^*$ when $\|\mathbf{x}^*\|_2 \neq 1$,

$$\begin{aligned}
\text{dist}(\mathbf{x}^{(0)}, \mathbf{x}^*) &\leq \text{dist}(\lambda_0 \widetilde{\mathbf{x}}, \|\mathbf{x}^*\|_2 \cdot \widetilde{\mathbf{x}}) + \text{dist}(\|\mathbf{x}^*\|_2 \cdot \widetilde{\mathbf{x}}, \mathbf{x}^*) \\
&\leq \max\{\sqrt{1.02 + 6\delta} - 1, 1 - \sqrt{0.99 - 5\delta}\} \cdot \|\mathbf{x}^*\|_2 + \widetilde{\tau} \cdot \|\mathbf{x}^*\|_2 \\
&\leq \frac{1}{10}\|\mathbf{x}^*\|_2,
\end{aligned}$$

where the last inequality can be easily achieved by choosing $\delta, \widetilde{\tau}$ to be sufficiently small. This completes the proof.

## A.2 Convergence Analysis

Here we consider the robust phase retrieval model with additional noise defined in (3.1), i.e., $y_i = |\mathbf{a}_i^\top \mathbf{x}^*| + \eta_i^* + \epsilon_i$ and analyze the stability guarantees for our proposed algorithm. The initialization for noisy model can be found in Section A.1. In this subsection, we focus on analyzing the gradient descent stage in Algorithm 1 assuming that the initialization stage has already generated an estimation which is close to the true signal. The final objective is to bound:

$$\text{dist}^2(\mathbf{x}^{(t+1)}, \mathbf{x}^*) = \text{dist}^2\Big(\mathbf{x}^{(t)} - \mu \cdot \nabla_\mathbf{x} L(\mathbf{x}^{(t)}, \boldsymbol{\eta}^{(t+1)}), \mathbf{x}^*\Big).$$

Notice that for the gradient update it is obvious that

$$-\mathbf{x}^{(t)} - \mu \nabla_\mathbf{x} L(-\mathbf{x}^{(t)}, \boldsymbol{\eta}^{(t+1)}) = -\Big\{\mathbf{x}^{(t)} - \mu \nabla_\mathbf{x} L(\mathbf{x}^{(t)}, \boldsymbol{\eta}^{(t+1)})\Big\},$$

hence by definition we have

$$\text{dist}^2\Big((-\mathbf{x}^{(t)}) - \mu \cdot \nabla_\mathbf{x} L(-\mathbf{x}^{(t)}, \boldsymbol{\eta}^{(t+1)}), \mathbf{x}^*\Big) = \text{dist}^2\Big(\mathbf{x}^{(t)} - \mu \cdot \nabla_\mathbf{x} L(\mathbf{x}^{(t)}, \boldsymbol{\eta}^{(t+1)}), \mathbf{x}^*\Big),$$

despite that the global phase function is unrecoverable. Thus we can get rid of the global phase term in distance function by letting $\mathbf{x}$ to be $e^{-j\phi(\mathbf{x})}\mathbf{x}$ for simplicity and directly set $\mathbf{h} = \mathbf{x}^{(t)} - \mathbf{x}^*$ for the analysis. Now we consider the (sub-)gradient of $L(\mathbf{x}, \boldsymbol{\eta})$ respect to $\mathbf{x}$ for the $t$−th iterate:

$$\begin{aligned}
\nabla_\mathbf{x} L(\mathbf{x}^{(t)}, \boldsymbol{\eta}^{(t+1)}) &= \frac{1}{m}\sum_{i=1}^m \Big(\big|\mathbf{a}_i^\top \mathbf{x}^{(t)}\big| + \eta_i^{(t+1)} - y_i\Big) \cdot \text{sgn}(\mathbf{a}_i^\top \mathbf{x}^{(t)}) \cdot \mathbf{a}_i \\
&= \frac{1}{m}\sum_{i=1}^m \big(\big|\mathbf{a}_i^\top \mathbf{x}^{(t)}\big| - \big|\mathbf{a}_i^\top \mathbf{x}^*\big|\big)\text{sgn}(\mathbf{a}_i^\top \mathbf{x}^{(t)})\mathbf{a}_i \\
&\quad - \frac{1}{m}\sum_{i=1}^m \big(\eta_i^* - \eta_i^{(t+1)} + \epsilon_i\big)\text{sgn}(\mathbf{a}_i^\top \mathbf{x}^{(t)})\mathbf{a}_i. \quad\quad (A.3)
\end{aligned}$$

First we try to further decompose the second term in the above equation. We again discuss the subset that each data sample belongs to. Denote the support for $\boldsymbol{\eta}^*$ as $\Omega^*$ and the support for $\boldsymbol{\eta}^{(t+1)}$ as $\Omega'$ and we have the following three cases:

- $i \in \Omega'$ : Since $i$ belongs to the support for $\boldsymbol{\eta}^{(t+1)}$, we have $\eta_i^{(t+1)} = y_i - |\mathbf{a}_i^\top \mathbf{x}^{(t)}|$. By model definition (3.1) we have $y_i = y_i^* + \eta_i^* + \epsilon_i$. It immediately implies that $\eta_i^* - \eta_i^{(t+1)} + \epsilon_i = |\mathbf{a}_i^\top \mathbf{x}^{(t)}| - |\mathbf{a}_i^\top \mathbf{x}^*|$.

- $i \notin \Omega', i \in \Omega^*$ : Since $i$ does not belong to the support for $\boldsymbol{\eta}^{(t+1)}$, $\eta_i^{(t+1)} = 0$. We have $\eta_i^* - \eta_i^{(t+1)} + \epsilon_i = \eta_i^* + \epsilon_i$.

- $i \notin \Omega', i \notin \Omega^*$ : Since $i$ does not belong to either the support for $\boldsymbol{\eta}^{(t+1)}$ or the support for $\boldsymbol{\eta}^*$, we immediately have $\eta_i^* - \eta_i^{(t+1)} + \epsilon_i = \epsilon_i$.

According to the above discussion, we can separate the summation over 1 to $m$ into three parts:

$$\frac{1}{m} \sum_{i=1}^m \left( \eta_i^* - \eta_i^{(t+1)} + \epsilon_i \right) \cdot \text{sgn}(\mathbf{a}_i^\top \mathbf{x}^{(t)}) \cdot \mathbf{a}_i$$

$$= \frac{1}{m} \left[ \sum_{i \in \Omega'} \left( |\mathbf{a}_i^\top \mathbf{x}^{(t)}| - |\mathbf{a}_i^\top \mathbf{x}^*| \right) + \sum_{i \in \Omega^* \setminus \Omega'} \left( \eta_i^* + \epsilon_i \right) + \sum_{i \notin (\Omega^* \cup \Omega')} \epsilon_i \right] \cdot \text{sgn}(\mathbf{a}_i^\top \mathbf{x}^{(t)}) \cdot \mathbf{a}_i$$

$$= \frac{1}{m} \sum_{i \in \Omega'} \left( |\mathbf{a}_i^\top \mathbf{x}^{(t)}| - |\mathbf{a}_i^\top \mathbf{x}^*| \right) \text{sgn}(\mathbf{a}_i^\top \mathbf{x}^{(t)}) \mathbf{a}_i + \frac{1}{m} \sum_{i \in \Omega^* \setminus \Omega'} \eta_i^* \cdot \text{sgn}(\mathbf{a}_i^\top \mathbf{x}^{(t)}) \mathbf{a}_i$$

$$+ \frac{1}{m} \sum_{i \notin \Omega'} \epsilon_i \cdot \text{sgn}(\mathbf{a}_i^\top \mathbf{x}^{(t)}) \mathbf{a}_i.$$

Note that the first term above can be merged with the first term in (A.3), thus we obtain

$$\nabla_{\mathbf{x}} L(\mathbf{x}^{(t)}, \boldsymbol{\eta}^{(t+1)}) = \underbrace{\frac{1}{m} \sum_{i \notin \Omega'} \left( |\mathbf{a}_i^\top \mathbf{x}^{(t)}| - |\mathbf{a}_i^\top \mathbf{x}^*| \right) \text{sgn}(\mathbf{a}_i^\top \mathbf{x}^{(t)}) \mathbf{a}_i}_{\textit{approximate gradient}}$$

$$- \frac{1}{m} \sum_{i \in \Omega^* \setminus \Omega'} \eta_i^* \cdot \text{sgn}(\mathbf{a}_i^\top \mathbf{x}^{(t)}) \mathbf{a}_i - \frac{1}{m} \sum_{i \notin \Omega'} \epsilon_i \cdot \text{sgn}(\mathbf{a}_i^\top \mathbf{x}^{(t)}) \mathbf{a}_i.$$

For convenience, we define the first term in the R.H.S of the above equality as the approximate gradient:

$$\mathbf{g}(\mathbf{x}^{(t)}, \boldsymbol{\eta}^{(t+1)}) = \frac{1}{m} \sum_{i \notin \Omega'} \left( |\mathbf{a}_i^\top \mathbf{x}^{(t)}| - |\mathbf{a}_i^\top \mathbf{x}^*| \right) \text{sgn}(\mathbf{a}_i^\top \mathbf{x}^{(t)}) \mathbf{a}_i.$$

Thus we have

$$\text{dist}(\mathbf{x}^{(t+1)}, \mathbf{x}^*) \leq \left\| \mathbf{x}^{(t)} - \mu \cdot \nabla_{\mathbf{x}} L(\mathbf{x}^{(t)}, \boldsymbol{\eta}^{(t+1)}) - \mathbf{x}^* \right\|_2$$
$$= \left\| \mathbf{x}^{(t)} - \mathbf{x}^* - \mu \cdot \mathbf{g}(\mathbf{x}^{(t)}, \boldsymbol{\eta}^{(t+1)}) \right\|_2 + \mu \left\| \nabla_{\mathbf{x}} L(\mathbf{x}^{(t)}, \boldsymbol{\eta}^{(t+1)}) - \mathbf{g}(\mathbf{x}^{(t)}, \boldsymbol{\eta}^{(t+1)}) \right\|_2$$
$$= \underbrace{\left\| \mathbf{h} - \mu \cdot \mathbf{g}(\mathbf{x}^{(t)}, \boldsymbol{\eta}^{(t+1)}) \right\|_2}_{\textit{approximate distance}} + \mu \underbrace{\left\| \nabla_{\mathbf{x}} L(\mathbf{x}^{(t)}, \boldsymbol{\eta}^{(t+1)}) - \mathbf{g}(\mathbf{x}^{(t)}, \boldsymbol{\eta}^{(t+1)}) \right\|_2}_{\textit{residual gradient error}}. \qquad (A.4)$$

Note that for the approximate distance term, by Lemma 6.3 we have
$$\left\|\mathbf{h} - \mu \cdot \mathbf{g}(\mathbf{x}^{(t)}, \boldsymbol{\eta}^{(t+1)})\right\|_2^2 = \left(1 - 2\mu(0.73 - 3\delta) + \mu^2(1+\delta)^2\right)\|\mathbf{h}\|_2^2. \tag{A.5}$$
For the residual gradient error term, note that we have
$$\left\|\nabla_{\mathbf{x}} L(\mathbf{x}^{(t)}, \boldsymbol{\eta}^{(t+1)}) - \mathbf{g}(\mathbf{x}^{(t)}, \boldsymbol{\eta}^{(t+1)})\right\|_2$$
$$= \left\|\frac{1}{m}\sum_{i \in \Omega^* \setminus \Omega'} \eta_i^* \cdot \mathrm{sgn}(\mathbf{a}_i^\top \mathbf{x}^{(t)})\mathbf{a}_i + \frac{1}{m}\sum_{i \notin \Omega'} \epsilon_i \cdot \mathrm{sgn}(\mathbf{a}_i^\top \mathbf{x}^{(t)})\mathbf{a}_i\right\|_2$$
$$\leq \left\|\frac{1}{m}\sum_{i \in \Omega^* \setminus \Omega'} \eta_i^* \cdot \mathrm{sgn}(\mathbf{a}_i^\top \mathbf{x}^{(t)})\mathbf{a}_i\right\|_2 + \left\|\frac{1}{m}\sum_{i \notin \Omega'} \epsilon_i \cdot \mathrm{sgn}(\mathbf{a}_i^\top \mathbf{x}^{(t)})\mathbf{a}_i\right\|_2. \tag{A.6}$$
By Lemma 6.4 and Lemma 6.5 we obtain
$$\left\|\frac{1}{m}\sum_{i \in \Omega^* \setminus \Omega'} \eta_i^* \cdot \mathrm{sgn}(\mathbf{a}_i^\top \mathbf{x}^{(t)})\mathbf{a}_i\right\|_2 \leq 0.02 \cdot \|\mathbf{h}\|_2 + 2\sqrt{\alpha(1+\delta)} \cdot \|\boldsymbol{\epsilon}\|_\infty, \tag{A.7}$$
and
$$\left\|\frac{1}{m}\sum_{i \notin \Omega'} \epsilon_i \cdot \mathrm{sgn}(\mathbf{a}_i^\top \mathbf{x}^{(t)})\mathbf{a}_i\right\|_2 \leq \sqrt{1+\delta} \cdot \|\boldsymbol{\epsilon}\|_\infty. \tag{A.8}$$
Submit (A.7), (A.8) back into (A.6) we have the bound for the residual gradient error term:
$$\left\|\nabla_{\mathbf{x}} L(\mathbf{x}^{(t)}, \boldsymbol{\eta}^{(t+1)}) - \mathbf{g}(\mathbf{x}^{(t)}, \boldsymbol{\eta}^{(t+1)})\right\|_2 \leq 0.02 \cdot \|\mathbf{h}\|_2 + 2\sqrt{\alpha(1+\delta)} \cdot \|\boldsymbol{\epsilon}\|_\infty + \sqrt{1+\delta} \cdot \|\boldsymbol{\epsilon}\|_\infty. \tag{A.9}$$
Combine the above result by submitting (A.9) and (A.5) into (A.4) we have
$$\mathrm{dist}(\mathbf{x}^{(t+1)}, \mathbf{x}^*) \leq \left\|\mathbf{h} - \mu \cdot \mathbf{g}(\mathbf{x}^{(t)}, \boldsymbol{\eta}^{(t+1)})\right\|_2 + \mu\left\|\nabla_{\mathbf{x}} L(\mathbf{x}^{(t)}, \boldsymbol{\eta}^{(t+1)}) - \mathbf{g}(\mathbf{x}^{(t)}, \boldsymbol{\eta}^{(t+1)})\right\|_2$$
$$\leq \left(\sqrt{1 - 2\mu(0.73 - 3\delta) + \mu^2(1+\delta)^2} + 2\mu\delta\right) \cdot \|\mathbf{h}\|_2 + \left(2\sqrt{\alpha} + 1\right)\mu\sqrt{1+\delta} \cdot \|\boldsymbol{\epsilon}\|_\infty$$
$$\leq (1 - \mu/2) \cdot \|\mathbf{h}\|_2 + c_4\mu \cdot \|\boldsymbol{\epsilon}\|_\infty,$$
where the last inequality holds when choosing $\delta$ and $\alpha$ to be sufficiently small provided $\mu \leq \mu_0$ with $\mu_0$ as a positive universal constant. Next we are going to show the convergence result based on mathematical induction. For the first step in gradient descent stage, we have
$$\mathrm{dist}(\mathbf{x}^{(1)}, \mathbf{x}^*) \leq (1 - \mu/2) \cdot \mathrm{dist}(\mathbf{x}^{(0)} - \mathbf{x}^*) + c_4\mu \cdot \|\boldsymbol{\epsilon}\|_\infty \leq \frac{1}{10}\|\mathbf{x}^*\|_2,$$
as long as $\delta$ is chosen to be sufficiently small. For the $(t+1)$-th iteration, suppose we have
$$\mathrm{dist}(\mathbf{x}^{(t)}, \mathbf{x}^*) \leq \frac{1}{10}\|\mathbf{x}^*\|_2,$$
then it follows that
$$\mathrm{dist}(\mathbf{x}^{(t+1)}, \mathbf{x}^*) \leq (1 - \mu/2) \cdot \mathrm{dist}(\mathbf{x}^{(t)} - \mathbf{x}^*) + c_4\mu \cdot \|\boldsymbol{\epsilon}\|_\infty \leq \frac{1}{10}\|\mathbf{x}^*\|_2,$$
as long as $\delta$ is chosen to be sufficiently small. Thus we proved that for all iterations we have
$$\mathrm{dist}(\mathbf{x}^{(t)}, \mathbf{x}^*) \leq \frac{1}{10}\|\mathbf{x}^*\|_2.$$
Given this fact, direct computation leads to the following conclusion:
$$\mathrm{dist}(\mathbf{x}^{(t)}, \mathbf{x}^*) \leq \frac{1}{10}\left(1 - \frac{\mu}{2}\right)^t \cdot \|\mathbf{x}^*\|_2 + c_4\mu \cdot \|\boldsymbol{\epsilon}\|_\infty.$$

# B Proof of Technical Lemmas in Section 6

## B.1 Proof of Lemma 6.1

*Proof.* Note that $y_i^* = |\mathbf{a}_i^\top \mathbf{x}^*| = |\sum_{j=1}^n a_{ij} x_j|$ where $a_{ij}$ follows $N(0,1)$ distribution. Apply Hoeffding type inequality (Lemma C.5) and union bound to all $i$ we have with probability at least $1 - 1/m$,

$$\|\mathbf{y}^*\|_\infty \leq \sqrt{c_0 \log(m)} \cdot \|\mathbf{x}^*\|_2,$$

where $c_0$ is a universal constant. It is easy to see $\|\boldsymbol{\epsilon}\|_\infty \leq \delta \|\mathbf{x}^*\|_2 \leq \sqrt{c_0 \log(m)} \cdot \|\mathbf{x}^*\|_2$ with small $\delta$. Thus we have

$$\|\mathbf{y}^* + \boldsymbol{\epsilon}\|_\infty \leq \|\mathbf{y}^*\|_\infty + \|\boldsymbol{\epsilon}\|_\infty \leq 2\sqrt{c_0 \log(m)} \cdot \|\mathbf{x}^*\|_2 \leq \sqrt{c_1 \log(m)} \cdot \|\mathbf{x}^*\|_2, \tag{B.1}$$

By definition we have

$$\lambda_0^2 = \frac{1}{m} \sum_{i=1}^m \widehat{y}_i^{\,2} = \frac{1}{m} \sum_{i=1}^m (\widehat{y}_i - y_i^* - \epsilon_i + y_i^* + \epsilon_i)^2$$

$$= \underbrace{\frac{1}{m} \sum_{i=1}^m (y_i^* + \epsilon_i)^2}_{I_1} + \underbrace{\frac{1}{m} \sum_{i=1}^m (\widehat{y}_i - y_i^* - \epsilon_i)^2}_{I_2} + \underbrace{\frac{2}{m} \sum_{i=1}^m (y_i^* + \epsilon_i)(\widehat{y}_i - y_i^* - \epsilon_i)}_{I_3}. \tag{B.2}$$

First we try to give the upper bound for $\lambda_0$. For term $I_1$, by lemma C.1 we have

$$\frac{1}{m} \sum_{i=1}^m (y_i^* + \epsilon_i)^2 = \frac{1}{m} \sum_{i=1}^m (y_i^*)^2 + \frac{2}{m} \sum_{i=1}^m y_i^* \cdot \epsilon_i + \frac{1}{m} \sum_{i=1}^m (\epsilon_i)^2$$

$$\leq \frac{1}{m} \|\mathbf{y}^*\|_2^2 + \frac{2}{m} \|\mathbf{y}^*\|_2 \cdot \|\boldsymbol{\epsilon}\|_2 + \frac{1}{m} \|\boldsymbol{\epsilon}\|_2^2$$

$$\leq (1 + \delta + 2\delta\sqrt{1+\delta} + \delta^2) \|\mathbf{x}^*\|_2^2 \leq (1 + 6\delta) \|\mathbf{x}^*\|_2^2. \tag{B.3}$$

where the first inequality follows from Cauchy-Schwarz inequality. For term $I_2$ we have

$$\frac{1}{m} \sum_{i=1}^m (\widehat{y}_i - y_i^* - \epsilon_i)^2 = \frac{1}{m} \|\widehat{\mathbf{y}} - \mathbf{y}^* - \boldsymbol{\epsilon}\|_2^2$$

$$\leq \alpha(\gamma + 1) \|\widehat{\mathbf{y}} - \mathbf{y}^* - \boldsymbol{\epsilon}\|_\infty^2 \leq 4\alpha(\gamma + 1) \|\mathbf{y}^* + \boldsymbol{\epsilon}\|_\infty^2,$$

where the first inequality is implied by Claim 1 in Section A.1 and the second inequality follows from Claim 2 in Section A.1. By plug in (B.1) and the condition on $\alpha$ we can further have

$$\frac{1}{m} \sum_{i=1}^m (\widehat{y}_i - y_i^* - \epsilon_i)^2 \leq 4c_1 \alpha(\gamma + 1) \log(m) \|\mathbf{x}^*\|_2^2 \leq c_2. \tag{B.4}$$

For term $I_3$, similarly we have

$$\frac{2}{m} \sum_{i=1}^m (y_i^* + \epsilon_i)(\widehat{y}_i - y_i^* - \epsilon_i) \leq \frac{2}{m} \|\mathbf{y}^* + \boldsymbol{\epsilon}\|_\infty \cdot \|\widehat{\mathbf{y}} - \mathbf{y}^* - \boldsymbol{\epsilon}\|_1 \leq 2\alpha(\gamma + 1) \|\mathbf{y}^* + \boldsymbol{\epsilon}\|_\infty \cdot \|\widehat{\mathbf{y}} - \mathbf{y}^* - \boldsymbol{\epsilon}\|_\infty$$

$$\leq 4\alpha(\gamma + 1) \|\mathbf{y}^* + \boldsymbol{\epsilon}\|_\infty^2 \leq 4c_1 \alpha(\gamma + 1) \log(m) \cdot \|\mathbf{x}^*\|_2^2$$

$$\leq c_2, \tag{B.5}$$

where the first inequality is due to Hölder inequality, the last inequality follows from the condition that $\alpha$ and the rest is due to the same reason as for term $I_2$. By submitting (B.3), (B.4), (B.5) into (B.2) and choose $c_2 = 0.01$ we have the following upper bound:

$$\lambda_0^2 \leq (1.02 + 6\delta)\|\mathbf{x}^*\|_2^2.$$

Next we try to give the lower bound for $\lambda_0$. For term $I_1$, by lemma C.1 we have

$$\begin{aligned}\frac{1}{m}\sum_{i=1}^m (y_i^* + \epsilon_i)^2 &= \frac{1}{m}\sum_{i=1}^m (y_i^*)^2 + \frac{2}{m}\sum_{i=1}^m y_i^* \cdot \epsilon_i + \frac{1}{m}\sum_{i=1}^m (\epsilon_i)^2 \\ &\geq \frac{1}{m}\|\mathbf{y}^*\|_2^2 - \frac{2}{m}\|\mathbf{y}^*\|_2 \cdot \|\boldsymbol{\epsilon}\|_2 \\ &\geq (1 - \delta - 2\delta\sqrt{1+\delta})\|\mathbf{x}^*\|_2^2 \leq (1 - 5\delta)\|\mathbf{x}^*\|_2^2.\end{aligned} \quad (B.6)$$

where the first inequality follows from Cauchy-Schwarz inequality. For term $I_2$ we obviously have

$$\frac{1}{m}\sum_{i=1}^m (\widehat{y}_i - y_i^* - \epsilon_i)^2 \geq 0. \quad (B.7)$$

For term $I_3$, note that

$$\begin{aligned}\frac{2}{m}\sum_{i=1}^m (y_i^* + \epsilon_i)(\widehat{y}_i - y_i^* - \epsilon_i) &\geq -\frac{2}{m}\left|\sum_{i=1}^m (y_i^* + \epsilon_i)(\widehat{y}_i - y_i^* - \epsilon_i)\right| \\ &\geq -\frac{2}{m}\|\mathbf{y}^* + \boldsymbol{\epsilon}\|_2 \cdot \|\widehat{\mathbf{y}} - \mathbf{y}^* - \boldsymbol{\epsilon}\|_2,\end{aligned}$$

where the second inequality follows from Cauchy-Schwarz inequality. Following the same route as in (B.5) we have

$$\frac{2}{m}\sum_{i=1}^m (y_i^* + \epsilon_i)(\widehat{y}_i - y_i^* - \epsilon_i) \geq -c_2. \quad (B.8)$$

By submitting (B.6), (B.7), (B.8) into (B.2) we have the following lower bound:

$$\lambda_0^2 \geq (0.99 - 7\delta)\|\mathbf{x}^*\|_2^2.$$

This completes the proof. □

### B.2 Proof of Lemma 6.2

*Proof.* Since $\widehat{\mathbf{y}} = \mathbf{y} - \boldsymbol{\eta}^{(0)}$ only contains elements that is not in the $\gamma\alpha m$-largest magnitude elements in $\mathbf{y}$, we can rewrite each $\widehat{y}_i$ as

$$\widehat{y}_i = y_i \cdot \mathbb{1}\{y_i \leq \theta_{1-\gamma\alpha}(\{y_i\})\},$$

where $\theta_p(\cdot)$ represents the $p$-quantile of the given samples. Since the observation has at most $\alpha$ fraction corrupted, it is easy to see that

$$\theta_{1-(\gamma+1)\alpha}(\{y_i^* + \epsilon_i\}) \leq \theta_{1-\gamma\alpha}(\{y_i\}) \leq \theta_{1-(\gamma-1)\alpha}(\{y_i^* + \epsilon_i\}).$$

Applying Lemma C.4 we further have

$$\theta_{1-(\gamma+1)\alpha}(\{y_i^*\}) - \|\boldsymbol{\epsilon}\|_\infty \leq \theta_{1-\gamma\alpha}(\{y_i\}) \leq \theta_{1-(\gamma-1)\alpha}(\{y_i^* + \epsilon_i\}) + \|\boldsymbol{\epsilon}\|_\infty. \tag{B.9}$$

Note that

$$y_i^* = |\mathbf{a}_i^\top \mathbf{x}^*|^2 = \left(\frac{\mathbf{a}_i^\top \mathbf{x}^*}{\|\mathbf{x}^*\|_2}\right)^2 \cdot \|\mathbf{x}^*\|_2^2 = \left(\frac{\mathbf{a}_i^\top \mathbf{x}^*}{\|\mathbf{x}^*\|_2}\right)^2.$$

Since $\mathbf{a}_i^\top \mathbf{x}^*/\|\mathbf{x}^*\|_2$ follows standard Gaussian distribution, therefore $y_i^*$ follows $\mathcal{X}_1^2$ distribution with its CDF denoted as $K$. By Lemma C.3, we have

$$\left|\theta_p(\{y_i^*\}) - \theta_p(K)\right| \leq \delta, \tag{B.10}$$

with probability at least $1 - 2\exp(-c_0 m \delta^2)$. Combine (B.9) with (B.10) and recall the fact that $\|\boldsymbol{\epsilon}\|_\infty \leq \delta \|\mathbf{x}^*\|_2$ we further have

$$\theta_{1-(\gamma+1)\alpha}(K) - 2\delta \leq \theta_{1-\gamma\alpha}(\{y_i\}) \leq \theta_{1-(\gamma-1)\alpha}(K) + 2\delta.$$

Consequently, we have

$$\mathbb{1}\{y_i \leq \theta_{1-(\gamma+1)\alpha}(K) - 2\delta\} \leq \mathbb{1}\{y_i \leq \theta_{1-\gamma\alpha}(\{y_i\})\} \leq \mathbb{1}\{y_i \leq \theta_{1-(\gamma-1)\alpha}(K) + 2\delta\}.$$

Recall in Algorithm 1 we define $\mathbf{Y}$ as:

$$\mathbf{Y} := \frac{1}{m}\sum_{i=1}^m \widehat{y}_i^2 \mathbf{a}_i \mathbf{a}_i^\top = \frac{1}{m}\sum_{i=1}^m y_i^2 \mathbf{a}_i \mathbf{a}_i^\top \cdot \mathbb{1}\{y_i \leq \theta_{1-\gamma\alpha}(\{y_i\})\}$$

$$= \frac{1}{m}\sum_{i\in\mathcal{C}} y_i^2 \mathbf{a}_i \mathbf{a}_i^\top \cdot \mathbb{1}\{y_i \leq \theta_{1-\gamma\alpha}(\{y_i\})\} + \frac{1}{m}\sum_{i\notin\mathcal{C}}(y_i^* + \epsilon_i)^2 \mathbf{a}_i \mathbf{a}_i^\top \cdot \mathbb{1}\{y_i \leq \theta_{1-\gamma\alpha}(\{y_i\})\},$$

where $\mathcal{C}$ denotes the true corruption set. We can upper and lower bound $\mathbf{Y}$ by

$$\mathbf{Y} \succ \mathbf{Y}_1 := \frac{1}{m}\sum_{i\notin\mathcal{C}} \mathbf{a}_i \mathbf{a}_i^\top \left[(\mathbf{a}_i^\top \mathbf{x}^*)^2 - \delta|\mathbf{a}_i^\top \mathbf{x}^*|\right] \cdot \mathbb{1}\{y_i \leq \theta_{1-(\gamma+1)\alpha}(K) - 2\delta\} - \frac{\delta^2}{m}\sum_{i\notin\mathcal{C}} \mathbf{a}_i \mathbf{a}_i^\top,$$

$$\mathbf{Y} \prec \mathbf{Y}_2 := \frac{1}{m}\sum_{i\notin\mathcal{C}} \mathbf{a}_i \mathbf{a}_i^\top \left[(\mathbf{a}_i^\top \mathbf{x}^*)^2 + \delta|\mathbf{a}_i^\top \mathbf{x}^*|\right] \cdot \mathbb{1}\{y_i \leq \theta_{1-(\gamma-1)\alpha}(K) + 2\delta\}$$

$$+ \frac{\delta^2}{m}\sum_{i\notin\mathcal{C}} \mathbf{a}_i \mathbf{a}_i^\top + \frac{1}{m}\sum_{i\in\mathcal{C}} \mathbf{a}_i \mathbf{a}_i^\top \cdot (\theta_{1-(\gamma-1)\alpha}(K) + 2\delta)^2.$$

Simple calculation yields that

$$\mathbb{E}[\mathbf{Y}_1] = (1-\alpha)\left(\beta_1 \mathbf{x}^* \mathbf{x}^{*\top} + \beta_2 \mathbf{I} - \delta^2 \mathbf{I}\right),$$
$$\mathbb{E}[\mathbf{Y}_2] = (1-\alpha)\left(\beta_3 \mathbf{x}^* \mathbf{x}^{*\top} + \beta_4 \mathbf{I} - \delta^2 \mathbf{I}\right) + \alpha(\theta_{1-(\gamma-1)\alpha}(K) + 2\delta)^2 \cdot \mathbf{I},$$

with

$$\beta_1 := \mathbb{E}\bigg[\big(|\xi|^4 - |\xi|^2\big)\mathbb{1}\big\{y_i \leq \theta_{1-(\gamma+1)\alpha}(K) - 2\delta\big\}\bigg] - \delta \cdot \mathbb{E}\bigg[\big(|\xi|^3 - |\xi|\big)\mathbb{1}\big\{y_i \leq \theta_{1-(\gamma+1)\alpha}(K) - 2\delta\big\}\bigg],$$

$$\beta_2 := \mathbb{E}\bigg[|\xi|^2\mathbb{1}\big\{y_i \leq \theta_{1-(\gamma+1)\alpha}(K) - 2\delta\big\}\bigg] - \delta \cdot \mathbb{E}\bigg[|\xi| \cdot \mathbb{1}\big\{y_i \leq \theta_{1-(\gamma+1)\alpha}(K) - 2\delta\big\}\bigg],$$

$$\beta_3 := \mathbb{E}\bigg[\big(|\xi|^4 - |\xi|^2\big)\mathbb{1}\big\{y_i \leq \theta_{1-(\gamma-1)\alpha}(K) + 2\delta\big\}\bigg] + \delta \cdot \mathbb{E}\bigg[\big(|\xi|^3 - |\xi|\big)\mathbb{1}\big\{y_i \leq \theta_{1-(\gamma-1)\alpha}(K) + 2\delta\big\}\bigg],$$

$$\beta_4 := \mathbb{E}\bigg[|\xi|^2\mathbb{1}\big\{y_i \leq \theta_{1-(\gamma-1)\alpha}(K) + 2\delta\big\}\bigg] + \delta \cdot \mathbb{E}\bigg[|\xi| \cdot \mathbb{1}\big\{y_i \leq \theta_{1-(\gamma-1)\alpha}(K) + 2\delta\big\}\bigg],$$

where $\xi \sim \mathcal{N}(0,1)$. Note that $\mathbf{a}_i \mathbf{a}_i^\top (\mathbf{a}_i^\top \mathbf{x}^*)^2 \mathbb{1}\{|\mathbf{a}_i^\top \mathbf{x}^*| \leq c\}$ can be seen as the outer product of two identical subGaussian vector $\boldsymbol{b}_i = \mathbf{a}_i(\mathbf{a}_i^\top \mathbf{x}^*)\mathbb{1}\{|\mathbf{a}_i^\top \mathbf{x}^*| \leq c\}$ and $\mathbf{a}_i \mathbf{a}_i^\top |\mathbf{a}_i^\top \mathbf{x}^*| \mathbb{1}\{|\mathbf{a}_i^\top \mathbf{x}^*| \leq c\}$ can be seen as the outer product of two identical subGaussian vector $\boldsymbol{b}'_i = \mathbf{a}_i\sqrt{|\mathbf{a}_i^\top \mathbf{x}^*|}\mathbb{1}\{|\mathbf{a}_i^\top \mathbf{x}^*| \leq c\}$. Therefore by applying results on random matries with non-isotropic subGaussian rows (equation (5.26) in Vershynin (2010)), we can conclude that

$$\|\mathbf{Y}_1 - \mathbb{E}[\mathbf{Y}_1]\|_2 \leq \tau, \qquad \|\mathbf{Y}_2 - \mathbb{E}[\mathbf{Y}_2]\|_2 \leq \tau,$$

with probability at least $1 - 4\exp(-c\tau^2 m)$, provided that $m/n$ exceeds some large constant. Further we have $\|\mathbb{E}[\mathbf{Y}_1] - \mathbb{E}[\mathbf{Y}_2]\|_2 \leq \tau$ for sufficiently small $\delta, \alpha$. Therefore we have

$$\|\mathbf{Y} - \mathbb{E}[\mathbf{Y}_1]\|_2 \leq 3\tau.$$

Now applying Davis-Kahan $\sin\Theta$ Theorem (Davis, 1963; Yu et al., 2015) we obtain

$$\sin\Theta(\widetilde{\mathbf{x}}, \mathbf{x}^*) \leq \frac{\|\mathbf{Y} - \mathbb{E}[\mathbf{Y}_1]\|_2}{\lambda_{gap}/2} \leq \frac{6\tau}{(1-\alpha)(\beta_1 + \beta_2 - \delta^2)},$$

where $\lambda_{gap}$ is the eigengap between the largest and the second largest eigenvalues of $\mathbb{E}[\mathbf{Y}_1]$, $\widetilde{\mathbf{x}}$ is the leading eigenvector for $\mathbf{Y}$ and $\Theta(\widetilde{\mathbf{x}}, \mathbf{x}^*)$ denotes the angle between $\widetilde{\mathbf{x}}$ and $\mathbf{x}^*$. We further have

$$\text{dist}(\widetilde{\mathbf{x}}, \mathbf{x}^*) \leq \sqrt{2}\sin\Theta(\widetilde{\mathbf{x}}, \mathbf{x}^*) \leq \frac{6\sqrt{2}\tau}{(1-\alpha)(\beta_1 + \beta_2 - \delta^2)} \leq \widetilde{\tau},$$

for a given $\widetilde{\tau}$ with sufficiently small $\tau, \delta, \alpha$. $\square$

### B.3  Proof of Lemma 6.3

*Proof.* Note that we have

$$\left\|\mathbf{h} - \mu \cdot \mathbf{g}(\mathbf{x}^{(t)}, \boldsymbol{\eta}^{(t+1)})\right\|_2^2 = \|\mathbf{h}\|_2^2 - 2\mu\langle \mathbf{g}(\mathbf{x}^{(t)}, \boldsymbol{\eta}^{(t+1)}), \mathbf{h}\rangle + \mu^2\|\mathbf{g}(\mathbf{x}^{(t)}, \boldsymbol{\eta}^{(t+1)})\|_2^2. \tag{B.11}$$

For the inner product term we have

$$\langle \mathbf{g}(\mathbf{x}^{(t)}, \boldsymbol{\eta}^{(t+1)}), \mathbf{h} \rangle = \frac{1}{m} \sum_{i \notin \Omega'} \left( \mathbf{a}_i^\top \mathbf{x}^{(t)} - (\mathbf{a}_i^\top \mathbf{x}^*) \cdot \mathrm{sgn}((\mathbf{a}_i^\top \mathbf{x}^{(t)}) \cdot (\mathbf{a}_i^\top \mathbf{x}^*)) \right) \cdot (\mathbf{a}_i^\top \mathbf{h})$$

$$= \frac{1}{m} \Big( \sum_{i \notin \Omega'} (\mathbf{a}_i^\top \mathbf{h})^2 + 2 \sum_{i \in \mathcal{S}} (\mathbf{a}_i^\top \mathbf{x}^*) \cdot (\mathbf{a}_i^\top \mathbf{h}) \Big)$$

$$\geq \frac{1}{m} \sum_{i \notin \Omega'} (\mathbf{a}_i^\top \mathbf{h})^2 - \frac{2}{m} \sum_{i \in \mathcal{S}} |(\mathbf{a}_i^\top \mathbf{x}^*) \cdot (\mathbf{a}_i^\top \mathbf{h})|$$

$$\geq \frac{1}{m} \sum_{i=1}^m (\mathbf{a}_i^\top \mathbf{h})^2 - \frac{1}{m} \sum_{i \in \Omega'} (\mathbf{a}_i^\top \mathbf{h})^2 - \frac{2}{m} \sum_{i \in \mathcal{S}'} |(\mathbf{a}_i^\top \mathbf{x}^*) \cdot (\mathbf{a}_i^\top \mathbf{h})|, \qquad (\text{B.12})$$

where $\mathcal{S} := \{i : (\mathbf{a}_i^\top \mathbf{x}^*) \cdot (\mathbf{a}_i^\top \mathbf{x}^{(t)}) < 0, i \notin \Omega'\}$ and $\mathcal{S}' := \{i : (\mathbf{a}_i^\top \mathbf{x}^*) \cdot (\mathbf{a}_i^\top \mathbf{x}^{(t)}) < 0, 1 \leq i \leq m\}$. Further we have with probability at least $1 - 2\exp(-c_1 \delta^2 m)$ that

$$\frac{2}{m} \sum_{i \in \mathcal{S}'} |(\mathbf{a}_i^\top \mathbf{x}^*) \cdot (\mathbf{a}_i^\top \mathbf{h})| \leq \frac{1}{m} \sum_{i=1}^m [(\mathbf{a}_i^\top \mathbf{x}^*)^2 + (\mathbf{a}_i^\top \mathbf{h})^2] \mathbb{1}\{(\mathbf{a}_i^\top \mathbf{x}^*) \cdot (\mathbf{a}_i^\top \mathbf{x}^{(t)}) < 0\}$$

$$\leq \frac{1}{m} \sum_{i=1}^m [(\mathbf{a}_i^\top \mathbf{x}^*)^2 + (\mathbf{a}_i^\top \mathbf{h})^2] \mathbb{1}\{(\mathbf{a}_i^\top \mathbf{x}^*)^2 + (\mathbf{a}_i^\top \mathbf{x}^*) \cdot (\mathbf{a}_i^\top \mathbf{h}) < 0\}$$

$$\leq \frac{1}{m} \sum_{i=1}^m [(\mathbf{a}_i^\top \mathbf{x}^*)^2 + (\mathbf{a}_i^\top \mathbf{h})^2] \mathbb{1}\{|\mathbf{a}_i^\top \mathbf{x}^*| \leq |\mathbf{a}_i^\top \mathbf{h}|\}$$

$$\leq \frac{2}{m} \sum_{i=1}^m (\mathbf{a}_i^\top \mathbf{x}^*)^2 \mathbb{1}\{|\mathbf{a}_i^\top \mathbf{x}^*| \leq |\mathbf{a}_i^\top \mathbf{h}|\}$$

$$\leq (0.26 + 2\delta) \|\mathbf{h}\|_2^2, \qquad (\text{B.13})$$

where the last inequality follows from Lemma 3 in Zhang and Liang (2016) (presented in Lemma C.2). Moreover, note that $|\Omega'| = \gamma \alpha m$. Apply Lemma C.5 and union bound to all $i$ we have $\max_i |\mathbf{a}_i^\top \mathbf{h}| \leq \sqrt{c_2 \log(m)} \cdot \|\mathbf{h}\|_2$ with probability at least $1 - 1/m$. Thus we obtain

$$\frac{1}{m} \sum_{i \in \Omega'} (\mathbf{a}_i^\top \mathbf{h})^2 \leq c_2 \gamma \alpha \cdot \log m \cdot \|\mathbf{h}\|_2^2 \leq c_3 \cdot \|\mathbf{h}\|_2^2, \qquad (\text{B.14})$$

where the last inequality due to that $\alpha \leq c_3/(c_2 \gamma \cdot \log m)$. Choose $c_3 = 0.01$ and combine (B.13) and (B.14) with (B.12) we have

$$\langle \mathbf{g}(\mathbf{x}^{(t)}, \boldsymbol{\eta}^{(t+1)}), \mathbf{h} \rangle \geq (0.73 - 3\delta) \|\mathbf{h}\|_2^2. \qquad (\text{B.15})$$

For the square term $\|\mathbf{g}(\mathbf{x}^{(t)}, \boldsymbol{\eta}^{(t+1)})\|_2^2$, denote each element of $\mathbf{v}$ as $v_i = (|\mathbf{a}_i^\top \mathbf{x}^{(t)}| - |\mathbf{a}_i^\top \mathbf{x}^*|) \cdot \mathrm{sgn}(\mathbf{a}_i^\top \mathbf{x}^{(t)})$ and $|(\Omega')^C|$ as $s$ we have

$$\|\mathbf{g}(\mathbf{x}^{(t)}, \boldsymbol{\eta}^{(t+1)})\|_2^2 = \left\| \frac{1}{m} \sum_{i \notin \Omega'} \left( |\mathbf{a}_i^\top \mathbf{x}^{(t)}| - |\mathbf{a}_i^\top \mathbf{x}^*| \right) \mathrm{sgn}(\mathbf{a}_i^\top \mathbf{x}^{(t)}) \mathbf{a}_i \right\|_2^2 = \frac{1}{m^2} \|\mathbf{A}_s^\top \mathbf{v}_s\|_2^2$$

$$\leq \frac{1}{m^2} \|\mathbf{A}_s\|_2^2 \cdot \|\mathbf{v}_s\|_2^2 \leq \frac{1}{m^2} \|\mathbf{A}\|_2^2 \cdot \|\mathbf{v}\|_2^2,$$

where $\mathbf{A}_s \in \mathbb{R}^{s \times n}$ is a matrix with each row being $\mathbf{a}_i^\top, i \notin \Omega'$, $\mathbf{v}_s \in \mathbb{R}^s$ is a vector with each element being $v_i, i \notin \Omega'$, $\mathbf{v} \in \mathbb{R}^m$ is a vector with each element being $v_i$ and the first inequality is due to Cauchy-Schwarz inequality. By Theorem 5.32 in Vershynin (2010) we have $\|\mathbf{A}\|_2 \leq \sqrt{(1+\delta)m}$ with probability at least $1 - 2\exp(-c_1\delta^2 m)$. Consider

$$|v_i|^2 = \left|(|\mathbf{a}_i^\top \mathbf{x}^{(t)}| - |\mathbf{a}_i^\top \mathbf{x}^*|) \cdot \text{sgn}(\mathbf{a}_i^\top \mathbf{x}^{(t)})\right|^2 \leq \left||\mathbf{a}_i^\top \mathbf{x}^{(t)}| - |\mathbf{a}_i^\top \mathbf{x}^*|\right|^2 \leq \left(\mathbf{a}_i^\top \mathbf{h}\right)^2.$$

By Lemma C.1 we have

$$\|\mathbf{v}\|_2^2 = \sum_{i=1}^m |v_i|^2 = \sum_{i=1}^m \left(\mathbf{a}_i^\top \mathbf{h}\right)^2 = \sum_{i=1}^m \mathbf{h}^\top \mathbf{a}_i \mathbf{a}_i^\top \mathbf{h} \leq (1+\delta)m\|\mathbf{h}\|_2^2.$$

Therefore combine the above results we obtain

$$\|\mathbf{g}(\mathbf{x}^{(t)}, \boldsymbol{\eta}^{(t+1)})\|_2^2 \leq (1+\delta)^2 \|\mathbf{h}\|_2^2. \tag{B.16}$$

Submit (B.16) and (B.15) back into (B.11) and let $\delta$ to be small enough we have

$$\left\|\mathbf{h} - \mu \cdot \mathbf{g}(\mathbf{x}^{(t)}, \boldsymbol{\eta}^{(t+1)})\right\|_2^2 = \left(1 - 2\mu(0.73 - 3\delta) + \mu^2(1+\delta)^2\right)\|\mathbf{h}\|_2^2.$$

This completes the proof. $\square$

### B.4 Proof of Lemma 6.4

*Proof.* Denote $u_i := \eta_i^* \cdot \text{sgn}(\mathbf{a}_i^\top \mathbf{x}^{(t)})$, we have

$$\left\|\frac{1}{m} \sum_{i \in \Omega^* \backslash \Omega'} \eta_i^* \cdot \text{sgn}(\mathbf{a}_i^\top \mathbf{x}^{(t)}) \mathbf{a}_i\right\|_2 = \left\|\frac{1}{m} \sum_{i \notin \Omega'} \eta_i^* \cdot \text{sgn}(\mathbf{a}_i^\top \mathbf{x}^{(t)}) \mathbf{a}_i\right\|_2 = \frac{1}{m}\|\mathbf{A}_s^\top \mathbf{u}_s\|_2$$

$$\leq \frac{1}{m}\|\mathbf{A}_s\|_2 \cdot \|\mathbf{u}_s\|_2 \leq \frac{1}{m}\|\mathbf{A}\|_2 \cdot \|\mathbf{u}_s\|_2, \tag{B.17}$$

where $\mathbf{A}_s \in \mathbb{R}^{s \times n}$ is a matrix with each row being $\mathbf{a}_i^\top, i \notin \Omega'$, $\mathbf{u}_s \in \mathbb{R}^s$ is a vector with each element being $u_i, i \notin \Omega'$, the first equality is due to the fact that $\eta_i^*$ is nonzero only when $i$ belongs to the support set $\Omega^*$ and the first inequality follows from Cauchy-Schwarz inequality. Again, by Theorem 5.32 in Vershynin (2010) we have $\|\mathbf{A}\|_2 \leq \sqrt{(1+\delta)m}$ with probability at least $1 - 2\exp(-c_1\delta^2 m)$. Further note that

$$\|\mathbf{u}_s\|_2^2 = \sum_{i \notin \Omega'} |u_i|^2 = \sum_{i \notin \Omega'} |\eta_i^*|^2 = \sum_{i \in \Omega^* \backslash \Omega'} |\eta_i^*|^2, \tag{B.18}$$

where the last equality is again due to the fact that $\eta_i^*$ is nonzero only when $i$ belongs to the support set $\Omega^*$. Note that when samples are from the support set $\Omega^* \setminus \Omega'$, it implies that $\left|y_i - |\mathbf{a}_i^\top \mathbf{x}^{(t)}|\right| \leq \left|(\mathbf{y} - \mathcal{A}(\mathbf{x}^{(t)}))^{(\gamma\alpha m)}\right|$ where $\left[\mathcal{A}(\mathbf{x}^{(t)})\right]_i = |\mathbf{a}_i^\top \mathbf{x}^{(t)}|$. Since $\boldsymbol{\eta}^*$ has at most $\alpha m$ nonzero entries, we claim that

$$|\eta_i^*| \leq 2\|\mathbf{y}^* - \mathcal{A}(\mathbf{x}^{(t)}) + \boldsymbol{\epsilon}\|_\infty. \tag{B.19}$$

The reason behind this is simple: if otherwise, we would have

$$\begin{aligned}
|y_i - |\mathbf{a}_i^\top \mathbf{x}^{(t)}|| &= |y_i^* + \eta_i^* + \epsilon_i - |\mathbf{a}_i^\top \mathbf{x}^{(t)}|| \\
&\geq \|\mathbf{y}^* - \mathcal{A}(\mathbf{x}^{(t)}) + \boldsymbol{\epsilon}\|_\infty \\
&= \|\mathbf{y} - \mathcal{A}(\mathbf{x}^{(t)}) - \boldsymbol{\eta}^*\|_\infty \\
&\geq \left|(\mathbf{y} - \mathcal{A}(\mathbf{x}^{(t)}))^{\gamma\alpha m}\right|,
\end{aligned}$$

where the equality is due to the model defined in (3.1) and the last inequality follows that $\boldsymbol{\eta}^*$ has at most $\alpha m$ nonzero entries. Specifically, suppose in the worst case, all $\alpha m$ entries' of $\boldsymbol{\eta}^*$ have large magnitude which would be hard thresholded in the first place, then $\gamma\alpha m$-largest element would be one of the elements in $\mathbf{y} - \mathcal{A}(\mathbf{x}^{(t)})$, and thus less than or equal to $\|\mathbf{y} - \mathcal{A}(\mathbf{x}^{(t)}) - \boldsymbol{\eta}^*\|_\infty$ which causes the contradiction. Thus combining (B.18) and (B.19) we have

$$\begin{aligned}
\|\mathbf{u}_s\|_2 &\leq 2\sqrt{\alpha m} \cdot \|\mathbf{y}^* - \mathcal{A}(\mathbf{x}^{(t)}) + \boldsymbol{\epsilon}\|_\infty \\
&\leq 2\sqrt{\alpha m} \cdot \|\mathbf{y}^* - \mathcal{A}(\mathbf{x}^{(t)})\|_\infty + 2\sqrt{\alpha m} \cdot \|\boldsymbol{\epsilon}\|_\infty,
\end{aligned} \tag{B.20}$$

where the last inequality is due to triangle inequality. Since by Hoeffding type inequality (Lemma C.5) we have

$$\|\mathbf{y}^* - \mathcal{A}(\mathbf{x}^{(t)})\|_\infty = \max_i ||\mathbf{a}_i^\top \mathbf{x}^{(t)}| - |\mathbf{a}_i^\top \mathbf{x}^*|| \leq \max_i |\mathbf{a}_i^\top \mathbf{h}| \leq \sqrt{c_0 \log m} \cdot \|\mathbf{h}\|_2,$$

(B.20) can be further written as

$$\|\mathbf{u}_s\|_2 \leq 2\sqrt{c_0 \alpha m \cdot \log m} \cdot \|\mathbf{h}\|_2 + 2\sqrt{\alpha m} \cdot \|\boldsymbol{\epsilon}\|_\infty.$$

Submit the above result back into (B.17) we have

$$\begin{aligned}
\left\|\frac{1}{m} \sum_{i \in \Omega^* \setminus \Omega'} \eta_i^* \cdot \operatorname{sgn}(\mathbf{a}_i^\top \mathbf{x}^{(t)}) \mathbf{a}_i\right\|_2 &\leq 2\sqrt{c_0 \alpha(1+\delta) \cdot \log m} \cdot \|\mathbf{h}\|_2 + 2\sqrt{\alpha(1+\delta)} \cdot \|\boldsymbol{\epsilon}\|_\infty \\
&\leq 2\sqrt{c_2} \cdot \|\mathbf{h}\|_2 + 2\sqrt{\alpha(1+\delta)} \cdot \|\boldsymbol{\epsilon}\|_\infty,
\end{aligned}$$

where the last inequality holds as long as $\alpha \leq c_2/(c_0 \log m)$. This completes the proof with $c_2 = 10^{-4}$. $\square$

## B.5 Proof of Lemma 6.5

*Proof of Lemma 6.5.* Denote $w_i = \epsilon_i \cdot \operatorname{sgn}(\mathbf{a}_i^\top \mathbf{x}^{(t)})$ and $s = |(\Omega')^C|$, we have

$$\begin{aligned}
\left\|\frac{1}{m} \sum_{i \notin \Omega'} \epsilon_i \cdot \operatorname{sgn}(\mathbf{a}_i^\top \mathbf{x}^{(t)}) \mathbf{a}_i\right\|_2 &= \frac{1}{m} \|\mathbf{A}_s^\top \mathbf{w}_s\|_2 \leq \frac{1}{m} \|\mathbf{A}_s\|_2 \cdot \|\mathbf{w}_s\|_2 \\
&\leq \frac{1}{m} \|\mathbf{A}\|_2 \cdot \|\mathbf{w}_s\|_2 \leq \sqrt{\frac{1+\delta}{m}} \cdot \|\boldsymbol{\epsilon}\|_2 \leq \sqrt{1+\delta} \cdot \|\boldsymbol{\epsilon}\|_\infty,
\end{aligned}$$

where $\mathbf{A}_s \in \mathbb{R}^{s \times n}$ is a matrix with each row being $\mathbf{a}_i^\top, i \notin \Omega'$, $\mathbf{w}_s \in \mathbb{R}^s$ is a vector with each element being $w_i, i \notin \Omega'$, the first inequality follows from Cauchy-Schwarz inequality and the third inequality is due to the fact that $\|\mathbf{A}\|_2 \leq \sqrt{(1+\delta)m}$ with probability at least $1 - 2\exp(-c_1\delta^2 m)$ (Theorem 5.32 in Vershynin (2010)).

$\square$

## C  Additional Auxiliary Lemmas

**Lemma C.1** (Lemma 3.1 in Candes et al. (2013))**.** For any $0 < \delta < 1$, $\mathbf{a}_i \sim N(\mathbf{0}, \mathbf{I}_{n \times n})$ independently, if $m > c_0 \delta^{-2} n$, then for all $\mathbf{h} \in \mathbb{R}^n$, with probability at least $1 - 2\exp(-c_1 \delta^2 m)$ we have

$$(1-\delta)\|\mathbf{h}\|_2^2 \leq \frac{1}{m}\sum_{i=1}^{m}(\mathbf{a}_i^\top \mathbf{h})^2 \leq (1+\delta)\|\mathbf{h}\|_2^2,$$

where $c_0, c_1$ are universal constants.

**Lemma C.2** (Lemma 3 in Zhang and Liang (2016))**.** For any $0 < \delta < 1$, if $m > c_0 \delta^{-2} n \log \delta^{-1}$, then for all $\mathbf{h} \in \mathbb{R}^n$ satisfying $\|\mathbf{h}\|_2 \leq \|\mathbf{x}^*\|_2/10$, with probability at least $1 - 2\exp(-c_1 \delta^2 m)$ we have

$$\frac{1}{m}\sum_{i=1}^{m}(\mathbf{a}_i^\top \mathbf{h})^2 \cdot \mathbb{1}\big\{(\mathbf{a}_i^\top \mathbf{x}^*)(\mathbf{a}_i^\top \mathbf{h}) < 0\big\} \leq (0.13 + \delta)\|\mathbf{h}\|_2^2,$$

where $c_0, c_1$ are universal constants.

**Lemma C.3** (Lemma 1 in Zhang et al. (2016a))**.** Suppose $F$ is a valid cumulative distribution function with continuous density function denote by $f$ and samples $\{X_i\}$ are i.i.d. drawn from $F$. Denote $\theta_p(F)$ as the $p$-quantile of $F$. If $l < f(\theta) < L$ for all $\theta$ in $\{\theta : |\theta - \theta_p| \leq \delta\}$, then

$$|\theta_p(\{X_i\}_{i=1}^m) - \theta_p(F)| \leq \delta$$

with probability at least $1 - 2\exp(-2m\delta^2 l^2)$.

**Lemma C.4** (Lemma 2 in Zhang et al. (2016a))**.** For vectors $\boldsymbol{X} = (X_1, .., X_n), \boldsymbol{Y} = (Y_1, .., Y_n)$, their order statistics satisfy

$$|X_{(k)} - Y_{(k)}| \leq \|\boldsymbol{X} - \boldsymbol{Y}\|_\infty$$

for all $k = 1, .., n$.

**Theorem C.5** (Proposition 5.10 in Vershynin (2010))**.** Suppose $X_1, X_2, \ldots, X_n$ are independent centered sub-Gaussian random variables, and $K = \max_i \|X_i\|_{\psi_2}$, then for every $\mathbf{a} = [a_1, a_2, \ldots, a_n]^\top \in \mathbb{R}^n$ and for every $t > 0$, we have

$$\mathbb{P}\bigg(\bigg|\sum_{i=1}^{n} a_i X_i\bigg| > t\bigg) \leq \exp\bigg(-\frac{Ct^2}{K^2 \|\mathbf{a}\|_2^2}\bigg),$$

where $C > 0$ is a constant.

## References

AGARWAL, A., NEGAHBAN, S. and WAINWRIGHT, M. J. (2012). Noisy matrix decomposition via convex relaxation: Optimal rates in high dimensions. *The Annals of Statistics* 1171–1197.

ANANDKUMAR, A., JAIN, P., SHI, Y. and NIRANJAN, U. (2015). Tensor vs matrix methods: Robust tensor decomposition under block sparse perturbations. *arXiv preprint* .

Bunk, O., Diaz, A., Pfeiffer, F., David, C., Schmitt, B., Satapathy, D. K. and van der Veen, J. F. (2007). Diffractive imaging for periodic samples: retrieving one-dimensional concentration profiles across microfluidic channels. *Acta Crystallographica Section A: Foundations of Crystallography* **63** 306–314.

Candès, E. J., Li, X., Ma, Y. and Wright, J. (2011). Robust principal component analysis? *Journal of the ACM (JACM)* **58** 11.

Candes, E. J., Li, X. and Soltanolkotabi, M. (2015a). Phase retrieval from coded diffraction patterns. *Applied and Computational Harmonic Analysis* **39** 277–299.

Candes, E. J., Li, X. and Soltanolkotabi, M. (2015b). Phase retrieval via wirtinger flow: Theory and algorithms. *IEEE Transactions on Information Theory* **61** 1985–2007.

Candes, E. J., Strohmer, T. and Voroninski, V. (2013). Phaselift: Exact and stable signal recovery from magnitude measurements via convex programming. *Communications on Pure and Applied Mathematics* **66** 1241–1274.

Chai, A., Moscoso, M. and Papanicolaou, G. (2010). Array imaging using intensity-only measurements. *Inverse Problems* **27** 015005.

Chandrasekaran, V., Sanghavi, S., Parrilo, P. A. and Willsky, A. S. (2011). Rank-sparsity incoherence for matrix decomposition. *SIAM Journal on Optimization* **21** 572–596.

Chen, Y. and Candes, E. (2015). Solving random quadratic systems of equations is nearly as easy as solving linear systems. In *Advances in Neural Information Processing Systems*.

Chen, Y., Caramanis, C. and Mannor, S. (2013). Robust sparse regression under adversarial corruption. In *ICML (3)*.

Cherapanamjeri, Y., Gupta, K. and Jain, P. (2016). Nearly-optimal robust matrix completion. *arXiv preprint arXiv:1606.07315* .

Davis, C. (1963). The rotation of eigenvectors by a perturbation. *Journal of Mathematical Analysis and Applications* **6** 159–173.

Fienup, J. R. (1978). Reconstruction of an object from the modulus of its fourier transform. *Optics letters* **3** 27–29.

Fienup, J. R. (1982). Phase retrieval algorithms: a comparison. *Applied optics* **21** 2758–2769.

Gerchberg, R. W. (1972). A practical algorithm for the determination of phase from image and diffraction plane pictures. *Optik* **35** 237.

Goldfarb, D. and Qin, Z. (2014). Robust low-rank tensor recovery: Models and algorithms. *SIAM Journal on Matrix Analysis and Applications* **35** 225–253.

Goldstein, T. and Studer, C. (2017). Convex phase retrieval without lifting via phasemax. In *International Conference on Machine Learning*.

Gu, Q., Gui, H. and Han, J. (2014). Robust tensor decomposition with gross corruption. In *Advances in Neural Information Processing Systems*.

Gu, Q., Wang, Z. and Liu, H. (2016). Low-rank and sparse structure pursuit via alternating minimization. In *Proceedings of the 19th International Conference on Artificial Intelligence and Statistics*.

Hand, P. (2017). Phaselift is robust to a constant fraction of arbitrary errors. *Applied and Computational Harmonic Analysis* **42** 550–562.

Hand, P. and Voroninski, V. (2016). Corruption robust phase retrieval via linear programming. *arXiv preprint arXiv:1612.03547* .

Harrison, R. W. (1993). Phase problem in crystallography. *JOSA A* **10** 1046–1055.

Hsu, D., Kakade, S. M. and Zhang, T. (2011). Robust matrix decomposition with sparse corruptions. *IEEE Transactions on Information Theory* **57** 7221–7234.

Huang, K., Eldar, Y. C. and Sidiropoulos, N. D. (2016). Phase retrieval from 1d fourier measurements: Convexity, uniqueness, and algorithms. *IEEE Transactions on Signal Processing* **64** 6105–6117.

Klopp, O., Lounici, K. and Tsybakov, A. B. (2014). Robust matrix completion. *Probability Theory and Related Fields* 1–42.

Kolte, R. and Özgür, A. (2016). Phase retrieval via incremental truncated wirtinger flow. *arXiv preprint arXiv:1606.03196* .

Miao, J., Charalambous, P., Kirz, J. and Sayre, D. (1999). Extending the methodology of x-ray crystallography to allow imaging of micrometre-sized non-crystalline specimens. *Nature* **400** 342–344.

Miao, J., Ishikawa, T., Shen, Q. and Earnest, T. (2008). Extending x-ray crystallography to allow the imaging of noncrystalline materials, cells, and single protein complexes. *Annu. Rev. Phys. Chem.* **59** 387–410.

Millane, R. P. (1990). Phase retrieval in crystallography and optics. *JOSA A* **7** 394–411.

Netrapalli, P., Jain, P. and Sanghavi, S. (2013). Phase retrieval using alternating minimization. In *Advances in Neural Information Processing Systems*.

Netrapalli, P., Niranjan, U., Sanghavi, S., Anandkumar, A. and Jain, P. (2014). Non-convex robust pca. In *Advances in Neural Information Processing Systems*.

Pardalos, P. M. and Vavasis, S. A. (1991). Quadratic programming with one negative eigenvalue is np-hard. *Journal of Global Optimization* **1** 15–22.

Sun, J., Qu, Q. and Wright, J. (2016). A geometric analysis of phase retrieval. In *Information Theory (ISIT), 2016 IEEE International Symposium on*. IEEE.

Vershynin, R. (2010). Introduction to the non-asymptotic analysis of random matrices. *arXiv preprint arXiv:1011.3027* .

Waldspurger, I., dâĂŹAspremont, A. and Mallat, S. (2015). Phase recovery, maxcut and complex semidefinite programming. *Mathematical Programming* **149** 47–81.

Wang, G., Giannakis, G. B. and Chen, J. (2016a). Solving large-scale systems of random quadratic equations via stochastic truncated amplitude flow. *arXiv preprint arXiv:1610.09540* .

Wang, G., Giannakis, G. B. and Eldar, Y. C. (2016b). Solving systems of random quadratic equations via truncated amplitude flow. *arXiv:1605.08285* .

Wei, K. (2015). Solving systems of phaseless equations via kaczmarz methods: A proof of concept study. *Inverse Problems* **31** 125008.

Yi, X., Park, D., Chen, Y. and Caramanis, C. (2016). Fast algorithms for robust pca via gradient descent. In *Advances in neural information processing systems*.

Yu, Y., Wang, T., Samworth, R. J. et al. (2015). A useful variant of the davis–kahan theorem for statisticians. *Biometrika* **102** 315–323.

Zhang, H., Chi, Y. and Liang, Y. (2016a). Provable non-convex phase retrieval with outliers: Median truncated wirtinger flow. In *Proceedings of The 33rd International Conference on Machine Learning*.

Zhang, H. and Liang, Y. (2016). Reshaped wirtinger flow for solving quadratic system of equations. In *Advances in Neural Information Processing Systems*.

Zhang, H., Zhou, Y., Liang, Y. and Chi, Y. (2016b). Reshaped wirtinger flow and incremental algorithm for solving quadratic system of equations. *arXiv preprint arXiv:1605.07719* .